\pdfoutput=1

\documentclass[11pt]{article}

\usepackage[]{ACL2023}

\usepackage{times}
\usepackage{latexsym}
\usepackage{tabularx}
\usepackage{booktabs} 
\usepackage{multirow} 
\usepackage{multicol}
\usepackage{subfigure}

\usepackage{enumitem}
\usepackage{makecell}
\usepackage{ltxtable}
\usepackage{listings}
\usepackage{comment}
\usepackage[utf8]{inputenc}
\usepackage{textgreek}
\usepackage[T1]{fontenc}
\usepackage{listings}
\usepackage{xcolor}

\usepackage{algorithm}
\usepackage{algpseudocode}
\usepackage{amsmath} 
\usepackage{graphicx}

\usepackage[utf8]{inputenc}
\usepackage{booktabs}
\usepackage{microtype}

\usepackage{inconsolata}
\usepackage{tikz}
\nolinenumbers

\newcommand{\circlednum}[1]{%
  \tikz[baseline=(char.base)]{
    \node[shape=circle,fill=black,text=white,inner sep=0.01pt] (char) {#1};}}

\newif\ifshowcomments
\showcommentstrue 
\ifshowcomments
    \newcommand{\jz}[1]{{\color{blue}[JZ: #1]}}
    \newcommand{\xt}[1]{{\color{pink}[XT: #1]}}
    \newcommand{\lx}[1]{{\color{red}[LX: #1]}}
    \newcommand{\sr}[1]{{\color{green}[SR: #1]}}
    \newcommand{\new}[1]{{\color{black}#1}}
\else
    \newcommand{\jz}[1]{}
    \newcommand{\xt}[1]{}
    \newcommand{\lx}[1]{}
    \newcommand{\sr}[1]{}
\fi

\title{\textsc{Meeting Delegate}: Benchmarking LLMs\\ on Attending Meetings on Our Behalf}

\author{Lingxiang Hu $^1$\footnotemark[1] \quad Shurun Yuan $^2$\footnotemark[1] \quad Xiaoting Qin $^3$ \quad Jue Zhang $^3$\footnotemark[2]  \quad Qingwei Lin $^3$ \\ \quad \textbf{Dongmei Zhang} $^3$ \quad \textbf{Saravan Rajmohan} $^3$ \quad \textbf{Qi Zhang} $^3$ \\
    $^1$Northeastern University, China \hspace{0.5em}
    $^2$Peking University, China \hspace{0.5em}
    $^3$Microsoft
    }

\begin{document}
\maketitle

\renewcommand{\thefootnote}{\fnsymbol{footnote}}
\footnotetext[1]{Equal contribution. Work is done during an internship at Microsoft.} 
\footnotetext[2]{Corresponding author.} 
\renewcommand{\thefootnote}{\arabic{footnote}}

\begin{abstract}

In contemporary workplaces, meetings are essential for exchanging ideas and ensuring team alignment but often face challenges such as time consumption, scheduling conflicts, and inefficient participation. Recent advancements in Large Language Models (LLMs) have demonstrated their strong capabilities in natural language generation and reasoning, prompting the question: {\it{can LLMs effectively delegate participants in meetings?}} To explore this, we develop a prototype LLM-powered meeting delegate system and create a comprehensive benchmark using real meeting transcripts. Our evaluation reveals that GPT-4/4o maintain balanced performance between active and cautious engagement strategies. In contrast, Gemini 1.5 Pro tends to be more cautious, while Gemini 1.5 Flash and Llama3-8B/70B display more active tendencies. Overall, about 60\% of responses address at least one key point from the ground-truth. However, improvements are needed to reduce irrelevant or repetitive content and enhance tolerance for transcription errors commonly found in real-world settings. Additionally, we implement the system in practical settings and collect real-world feedback from demos. Our findings underscore the potential and challenges of utilizing LLMs as meeting delegates, offering valuable insights into their practical application for alleviating the burden of meetings.


\end{abstract}

\section{Introduction}

Nowadays, the nature of work has increasingly become more collaborative~\cite{mckinseyfutureofwork}, with meetings becoming an essential component~\cite{Spataromeeting} to facilitate the exchange of ideas and information, fostering innovation and ensuring alignment among team members. 

Attending meetings, however, poses notable difficulties. Firstly, the rapid increase in the number of meetings can consume a substantial amount of time, diverting attention from core tasks and reducing overall productivity \cite{Perlowmeeting2017, Kostmeeting2020}. Secondly, scheduling conflicts often arise when multiple meetings are double-booked, forcing participants to prioritize or miss valuable discussions altogether. Thirdly, not all meetings require full attendance; participants may only need to contribute to specific topics, leading to inefficiencies when attendees are required for entire duration.

In this study, we investigate the feasibility of developing a meeting delegate system to represent individuals in meetings. This concept is becoming increasingly viable with the advancement of Large Language Models (LLMs). These LLMs, renowned for their remarkable capabilities in natural language understanding and generation~\cite{ouyang2022training, openai2023gpt4, google2024gemini}, demonstrate potential to comprehend meeting context, participate in dynamic conversations, and provide informed responses. 

Developing LLM-powered meeting delegate systems faces several challenges. Firstly, such systems must navigate complex, context-rich conversations involving multiple participants, requiring them to discern opportune moments for engagement and restraint. Secondly, human conversations often contain ambiguities and uncertainties, such as queries directed ambiguously or pronunciation-related ambiguities, which challenge the system's ability to respond effectively. Thirdly, ensuring user privacy is crucial to prevent over-sharing of information and safeguard the user's personal image. Finally, these systems must operate in real-time, necessitating low-latency responsiveness.

In this work we aim to develop a prototype of an LLM-powered meeting delegate system to address the above challenges, focusing initially on the first two while leaving the last two in the future work. To assess its effectiveness across various LLMs, we conduct real-world testing in a few demo scenarios and construct an evaluation dataset from real meeting transcripts. In contrast to recent studies that emphasize the facilitator role in meeting engagement~\cite{mao2024muca}, our work concentrates on the participant role, which is more prevalent and distinct from that of the facilitator.

Our evaluation reveals that GPT-4/4o maintain balanced performance between active and cautious engagement strategies, while Gemini 1.5 Pro is more cautious, and Gemini 1.5 Flash and Llama3-8B/70B are more active. \new{Overall, 60\% of responses address at least one main point from the ground-truth, showing the promise of adopting LLM-powered meeting delegates, while improvements are needed, such as to enhance tolerance for transcription errors.} 

Our contributions are summarized as follows:
\begin{itemize}[noitemsep, left=0pt]
    \item We develop a prototype of an LLM-powered meeting delegate system designed to participate in meetings on our behalf, with a particular focus on the role of the participant.
    \item We create a comprehensive benchmark based on real meeting transcripts, covering four common scenarios: Explicit Cue, Implicit Cue, Chime In, and Keep Silence. \new{We plan to release the benchmark dataset with the paper.}
    \item We evaluate the performance of popular LLMs through some demo scenarios and a rigorous assessment using the benchmark. This includes an ablation study on the impact of transcription errors commonly encountered in practice.
\end{itemize}

\section{Related Work}

\noindent{\textbf{Language Model Applications in Meetings.}} Considerable research has been dedicated to the summarization of meetings~\cite{Zhong2021qmsum} and other real-life dialogues~\cite{Mehdad2014summary, Tuggener2021summarysurvey}. In the context of meetings, key tasks include meeting transcript summarization and action item identification~\cite{Cohen2021action}. MeetingQA~\cite{Prasad2023meetingqa} investigated Q\&A tasks based on meeting transcripts, highlighting the challenges faced by models such as RoBERTa in handling real-world meeting data. Recent advancements in LLMs have opened new avenues for enhancing these tasks. For instance, an LLM-based meeting recap system~\cite{Asthana2023recap} has demonstrated effectiveness in generating accurate and coherent summaries and action items.



\noindent{\textbf{Facilitator in Multi-Participant Chat.}} MUCA \cite{mao2024muca} presents a framework that leverages LLMs to facilitate group chats by simulating users, demonstrating notable effectiveness in goal-oriented conversations. Similarly, approaches like GPT-4o demo for meetings~\cite{openaivoice} are designed to serve as facilitators in group discussions. While these studies underscore LLMs’ capabilities in managing group chats, they primarily focus on LLMs guiding the meeting process rather than representing individuals with different roles.


\noindent{\textbf{Role-Playing with LLMs: Characters and Digital Twins.}}
Role-play prompting~\cite{kong2024roleplay} has been shown to be a more effective trigger for the chain-of-thought process in LLMs. Additionally, efforts to simulate famous personalities~\cite{shao2023character, Sun2024persona} have garnered interest, leading to research on maintaining character consistency and studying social interactions within agent-based group chat environments. Recently, Reid Hoffman~\cite{ReidHoffman} showcased an interview between himself and his digital twin built on GPT-4. Although this demonstration highlighted the potential of digital representations, it was confined to one-on-one interactions, leaving the complexities of group discussions unexplored. Unlike previous work, our work focus on LLMs as meeting participant delegates, delivering targeted engagement tailored to multi-participant, meeting-specific objectives. Our comprehensive evaluation and real-world deployment further demonstrate the system's potential to significantly reduce the burden of meetings on individuals, thereby advancing the application of LLMs in professional environments.



\section{LLM-based Meeting Delegate System}

\begin{figure}[htb!]
  \centering
  \includegraphics[width=\columnwidth]{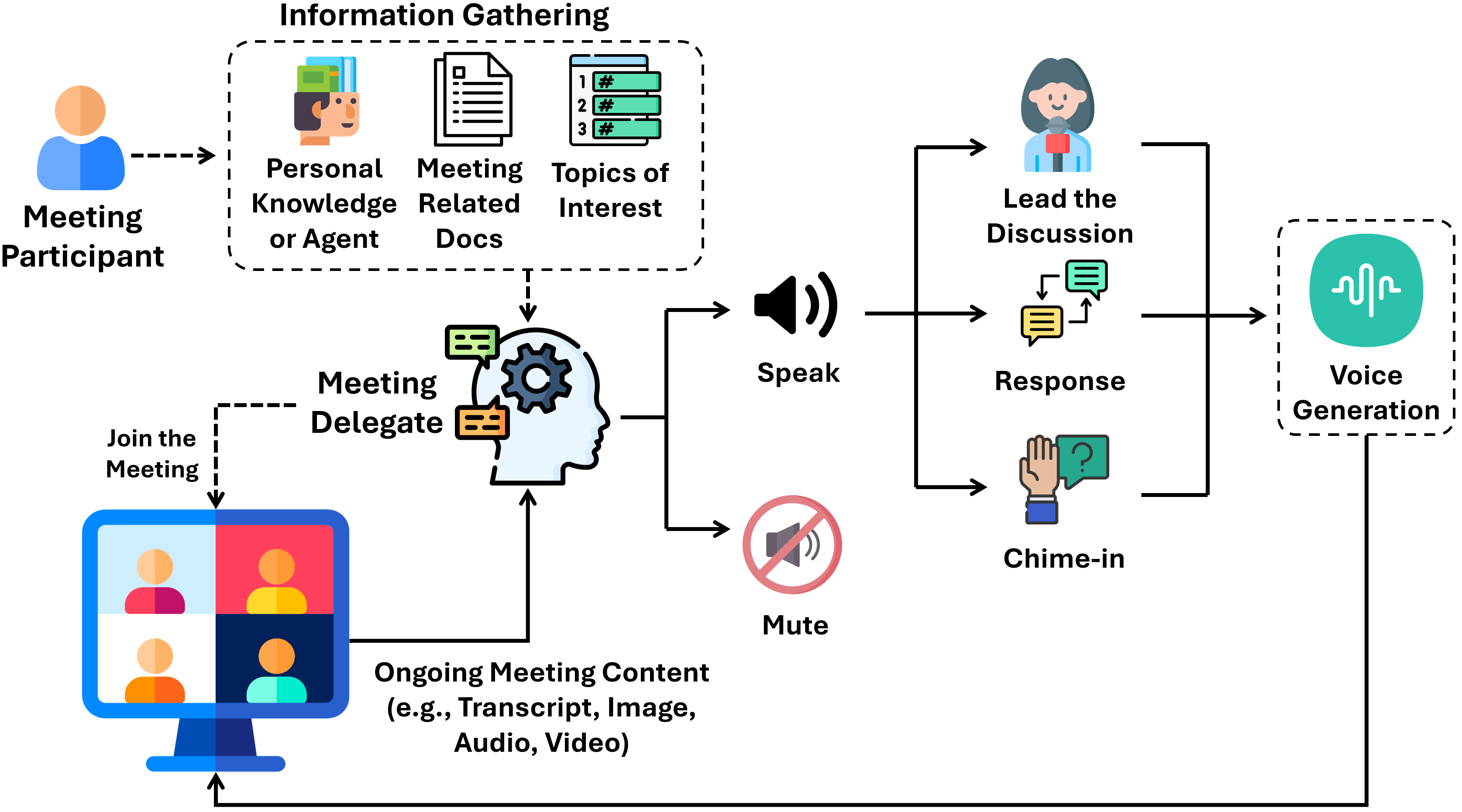}
  \caption{Architecture of the meeting delegate system.}
  \label{fig:shadowclone_diagram}
\end{figure}

Figure~\ref{fig:shadowclone_diagram} illustrates the architecture of our proposed meeting delegate system, which comprises three main components: 
\begin{itemize}[noitemsep, left=0pt]
    \item {\it\textbf{Information Gathering}}: This component shown on the top-left collects meeting-related information to assist LLMs in participating in meetings. Users can manually provide topics of interest, background knowledge, and shareable materials prior to the meeting. Alternatively, if the user has a personal knowledge base or an intelligent personal assistant/agent, the system can query them in real-time, provided latency is manageable.
    \item {\it\textbf{Meeting Engagement}}: The Meeting Engagement module actively monitors the meeting's status and uses LLMs to determine the appropriate timing and content for engagement. Engagement evaluation occurs after each participant's utterance, using in-context learning methods. The prompt for evaluation includes general instructions, user-provided meeting information, and the ongoing meeting context (see Table~\ref{tab:prompt_generate_response} in the Appendix for details). While various contextual data (e.g., transcript, screen sharing, audio) can be utilized, this work focuses on transcripts obtained via meeting software or speech-to-text tools. Figure~\ref{fig:shadowclone_diagram} shows the three common response types: leading the discussion, responding to others, and chiming in. This work concentrates on the latter two, emphasizing the participant's role.
    \item {\it\textbf{Voice Generation}}: After deciding on the content to be spoken, the Voice Generation module shown on the right produces a voice response mimicking the user's voice using text-to-speech (TTS) technology~\cite{qin2023openvoice}. To minimize latency, the system employs streaming modes for both LLM API calls and TTS. 

\end{itemize}

\begin{figure*}[htb!]
  \centering
  \includegraphics[width=\textwidth]{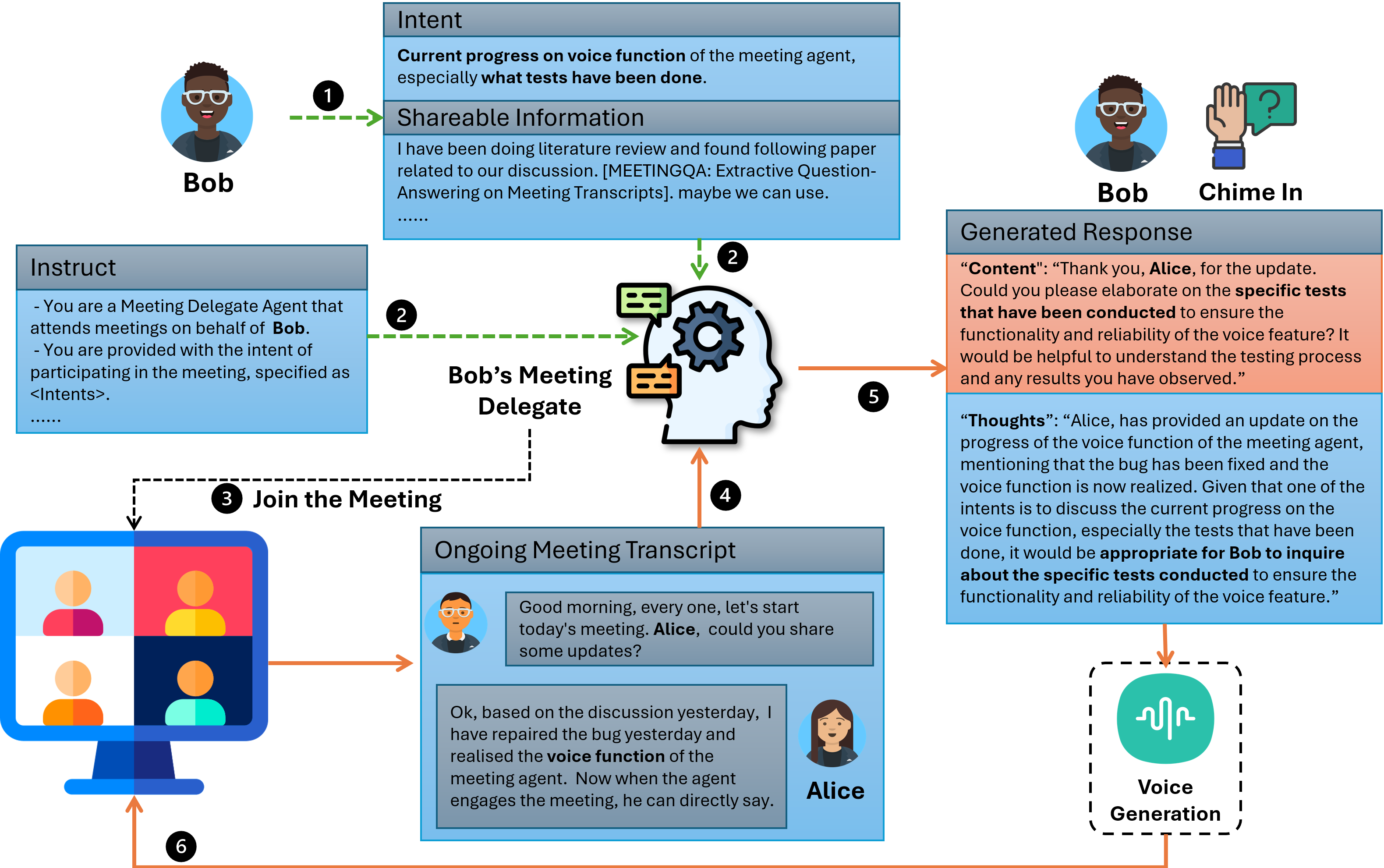}
  \caption{Workflow of an LLM-powered meeting delegate system. The process involves user input of meeting intent and shareable information prior to the meeting, real-time participation based on meeting transcripts, and response generation aligned with prompted instructions and meeting objectives.}
  \label{fig:demo}
\end{figure*}

\new{We implemented a prototype of the above system on a widely-used meeting platform and conducted several demo case studies.\footnote{Omitting the platform name for anonymity.} Detailed insights and lessons learned from these real-world applications will be presented in Section~\ref{sec:app_in_practice}. As an illustration, we present a real demo case in Figure~\ref{fig:demo}. In this example, Bob uses his Meeting Delegate to participate in a meeting with Alice and others. Before the meeting, Bob provides topics of interest and relevant shareable information to the Meeting Delegate \circlednum{1}. This information, along with instructions, forms the prompt for the Meeting Delegate \circlednum{2}. The delegate then joins the meeting \circlednum{3} and determines, based on the ongoing meeting transcript, whether to engage \circlednum{4}. During the meeting, Alice discusses updates on the voice function, which aligns with Bob's goal to learn about its progress. The Meeting Delegate then chimes in \circlednum{5}, generating a text-based response (converted to speech \circlednum{6}), asking for more details, thus achieving Bob's objectives and engaging in the conversation.
}





\section{Benchmark Dataset}
While the proposed meeting delegate system demonstrates potential in a few sample demonstrations, more systematic and quantitative evaluation in diverse contexts is needed. Our evaluation goals are to determine whether the system can appropriately time its interventions and generate relevant spoken content. No existing benchmark datasets meet these objectives, prompting us to create one. 

\subsection{Dataset Construction}

Our dataset construction strategy involves using real meeting transcripts and generating test cases by taking ``snapshots'' from these transcripts. A ``snapshot'' is defined as a truncation of the transcript after a participant's utterance. Then, by comparing the generated response according to this snapshot with the actual responses in the real script, we can determine how well the system performs. An illustration of this process in given in Figure~\ref{fig:dataset_construction}.

Our base meeting transcripts are taken from the ELITR Minuting Corpus~\cite{Nedoluzhko2022ELITRMC}, comprising de-identified project meeting transcripts in English and Czech. 61 English meeting transcripts are used and the test cases are constructed as follows. \new{Motivated by promising results from LLM annotation~\cite{llmannotation}, we also leverages LLMs for preparing this dataset while conducting manual verification for quality assurance.} Specifically, we first employ GPT-4 to progressively analyze each participant's utterances by taking a ``sliding window'' on the original meeting transcript. This is to capture their meeting intents and the information that they can share during the meeting, serving as the critical input to the Meeting Engagement module for response generation. The shareable meeting information contains pairs of <Context> and <Information>, with <Context> specifying under which context the points in <Information> can be shared. Details of this intent and contextual information extraction prompt can be found in Table~\ref{tab:prompt_context_extraction} in the Appendix.

Next, we extract suitable snapshots from the transcripts as test cases. For each participant (excluding facilitators), we identify their utterances and use the preceding transcript as the ongoing meeting context. The ground-truth response is determined by considering \textit{several} subsequent utterances. This extraction process leverages GPT-4 (prompt in Table~\ref{tab:prompt_case_extraction}), which classifies the meeting scenes (Explicit Cue, Implicit Cue, and Chime In) and selects the necessary utterances to form the ground-truth response, recognizing that a user's response may span multiple subsequent utterances. To ensure accuracy, the extracted cases are manually verified by two authors. As the extracted test cases closely match real transcripts, we refer them as the \textbf{Matched Dataset}.


To evaluate the meeting delegate's ability to remain silent when inappropriate to speak, we construct a \textbf{Mismatched Dataset} from the Matched Dataset. We take Explicit Cue and Implicit Cue test cases and replace the principal who needs to respond with another participant not involved in the current conversation. The intents and shareable meeting information are accordingly replaced, and the ground-truth is set to be empty. The delegate representing the new principal is expected to remain silent when presented with these transcripts.

Lastly, we construct a \textbf{Noisy Name Dataset} for our ablation study, addressing the fact that meeting transcribing systems often introduce noise affecting the meeting delegate's performance. This issue is particularly significant for recognizing names, which are crucial in Explicit Cue cases. For example, the Chinese name ``Jisen'' might be transcribed as ``Jason''. In our construction, we modify the Explicit Cue cases by replacing de-identified names with real-world names and substituting the principal's name in the final utterance with a phonetically similar word to simulate transcription errors.

\subsection{Evaluation Metric} 

In our evaluation, we generate responses using LLMs with the same prompt as in our prototype. These responses are assessed using two categories of metrics: \textbf{Response Rate / Silence Rate}, which determines whether a response is generated, and quality-related metrics, \textbf{Recall} and \textbf{Attribution}. 

The Recall metric evaluates if the generated response includes key points present in the ground-truth response. We define two recall rates: ``loose'' recall rate, which is 1 if at least one main point from the ground-truth is mentioned and 0 otherwise; and ``strict'' recall rate, which measures the percentage of main points from the ground-truth included in the generated response.

Attribution assesses the origin of the main points in the generated response, classifying them into four categories: the expected ground-truth response (Expected Response), contextual information not present in the ground-truth (Contextual Information), previous transcript content (Previous Transcript), and hallucinated texts (Hallucination).

We leverage LLMs for main point extraction and their semantic comparison. Specifically, in the Recall phase, GPT-4 is employed to assess how well the LLM-generated responses match key points from the ground-truth response set, using the prompt provided in Table~\ref{tab:prompt_evaluation}. In the Attribution phase, GPT-4 Turbo is used to trace and evaluate the origin of specific points in the responses, with the prompt provided in Table~\ref{tab:prompt_attribution}. \new{Through manual validation of 30 randomly sampled cases, we found that LLMs achieved an average of 93.3\% accuracy on these Recall and Attribution evaluation tasks, supporting their use in our experiments.
}

\end{itemize}
\end{comment}

\subsection{Dataset Statistics}

\begin{figure}[htb!]
  \centering
  \includegraphics[scale=0.36]{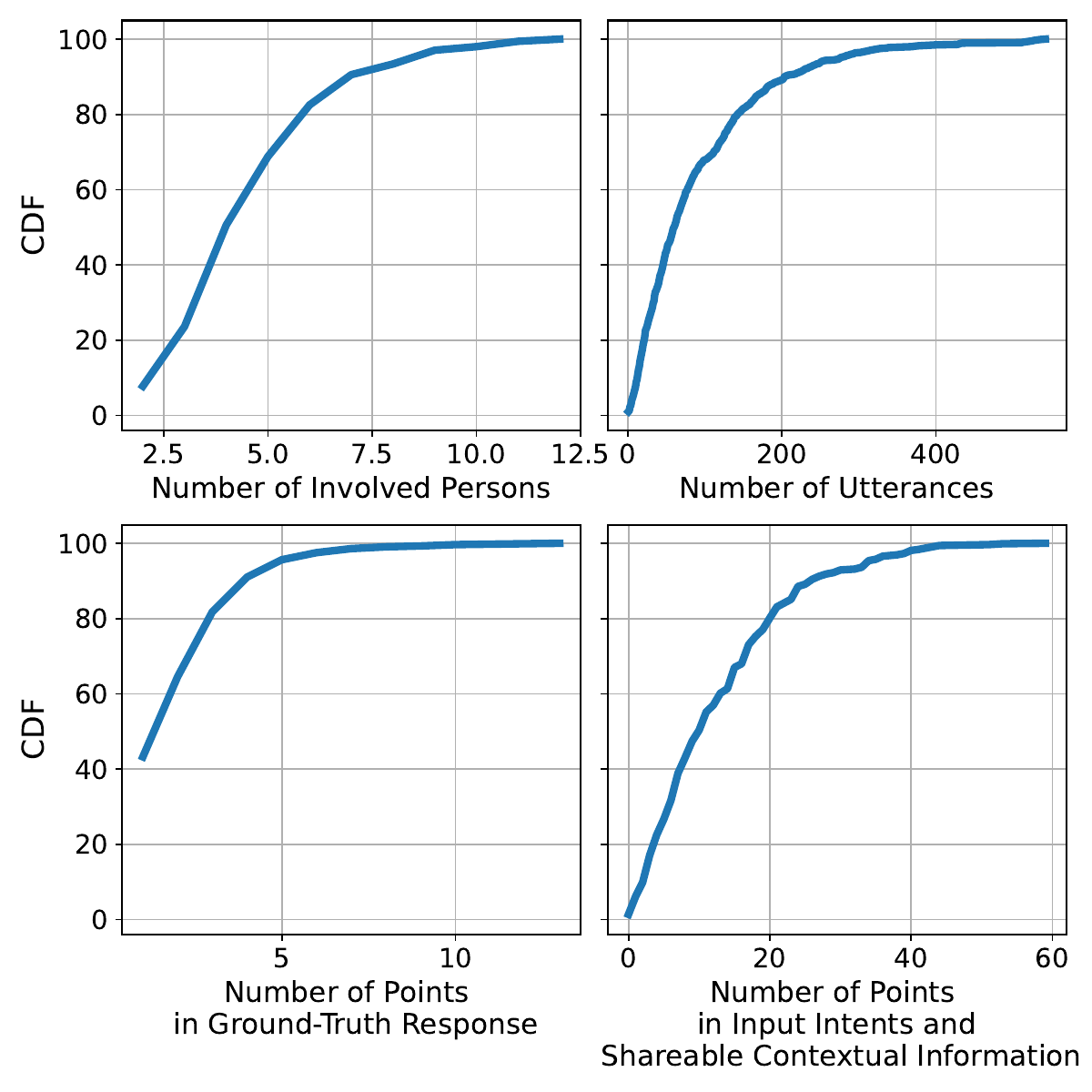}
  \caption{Data statistics of the Matched Dataset.}
  \label{fig:meta_data}

\end{figure}

From the 61 original meeting transcripts, we extract 846 test cases for Matched Dataset, in which 54.5\% belongs to Implicit Cue, followed by 30.9\% for Explicit Cue and 14.7\% for Chime In. The numbers of test cases for Mismatched Dataset and Noisy Name Dataset are 294 and 122, respectively.

For Matched Dataset, we present various data statistics in Figure~\ref{fig:meta_data}. Over 50\% of test cases involve more than four participants and contain transcripts exceeding 50 utterances, highlighting the dataset's complexity and the involvement of multiple individuals. Additionally, approximately 40\% of test cases include at least two main points in the ground-truth response, and in more than 50\% of cases, participants contribute over ten main points. This indicates a substantial level of detail and interaction within the meetings, suggesting that the dataset captures rich and multifaceted discussions.

\section{Experiment}
\label{sec:experiment}

\noindent\textbf{Setup.} In our experiment, we utilize three prominent series of LLMs: the GPT series (GPT-3.5-Turbo, GPT-4, GPT-4o)~\cite{openaimodel}, the Gemini series (Gemini 1.5 Flash, Gemini 1.5 Pro)~\cite{geminimodels} and the Llama series (Llama3-8B, Llama3-70B)~\cite{llama}. For all LLMs\footnote{Exact model versions can be found in Table~\ref{tab:model_use}.}, we set the temperature to 0 and use the default API settings for other parameters. Note that, due to model context window restriction, we remove test cases that exceed the 8K context window for Llama3 models (56.3\% kept) and those exceeding the 16K context window for GPT-3.5-Turbo (94.3\% kept), while keeping all for the other LLMs.

\begin{figure}[htb!]
  \centering
  \includegraphics[scale=0.46]{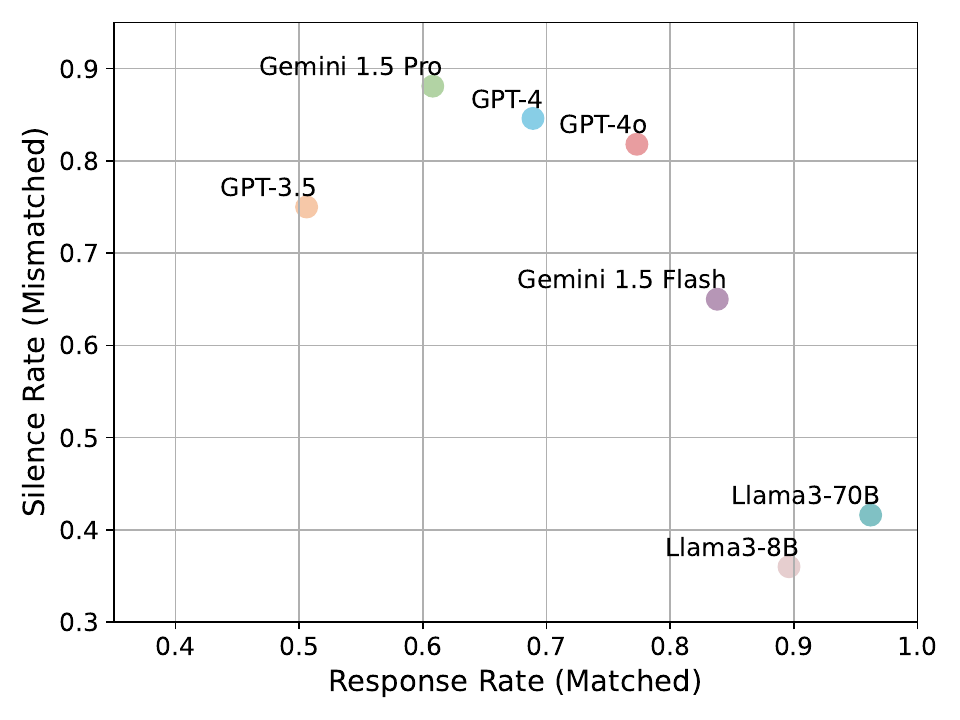}
  \caption{Response Rate on Matched Dataset vs. Silence Rate on Mismatched Dataset.}
  \label{fig:response_silence}
\end{figure}

\noindent\textbf{Response Rate Analysis.} The Response and Silence Rates of the studied LLMs are obtained for Matched and Mismatched Datasets, respectively. Summarized results are presented in Figure~\ref{fig:response_silence}, with further details \new{(e.g., breaking down to different meeting scenes)} provided in Tables~\ref{tab:Response_rate} and \ref{tab:Mismatch_Response_rate} in the Appendix. Overall, GPT-4 and GPT-4o demonstrated \textit{balanced} performance, with Response/Silence Rates between 0.7 and 0.8. Among the Gemini series models, Gemini 1.5 Pro achieved the highest Silence Rate of approximately 0.9, coupled with a low Response Rate, indicating a \textit{cautious} engagement strategy. In contrast, the smaller Gemini 1.5 Flash model and the Llama series exhibited higher activity levels, suggesting a more \textit{proactive} engagement approach; however, this also led to a tendency to engage when they should remain silent. These patterns persisted when all LLMs are tested using the same subset of cases as the Llama series.


\begin{figure}[htb!]
  \centering
  \includegraphics[scale=0.17]{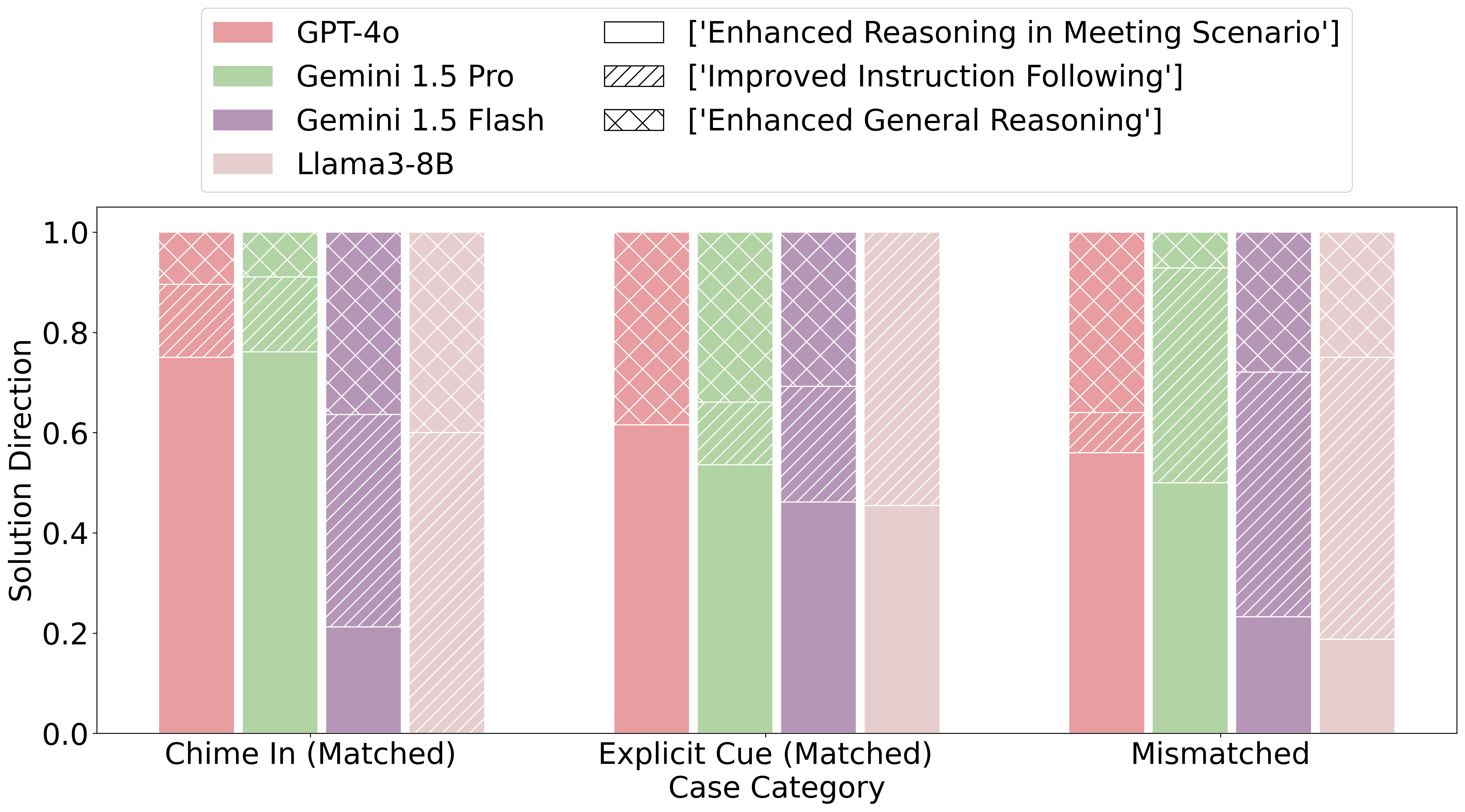}
  \caption{Solution directions from error analysis of bad cases in Response (Silence) Rate for Matched and Mismatched Datasets.}
  \label{fig:solution}
\end{figure}

To uncover the underlying causes of failures, we conduct an in-depth analysis of all failure cases in representative models: GPT-4o and Gemini 1.5 Pro for state-of-the-art LLMs, and Gemini 1.5 Flash and Llama3-8B representing more lightweight models. \new{We manually analyze and categorize all error types, proposing corresponding directions for improvement. For instance, in the "Explicit Cue" scenario within the Matched Dataset, the meeting delegate may correctly identify the cue but fail to respond, indicating a need for enhanced reasoning capabilities in meeting contexts. Detailed analysis can be found in Table~\ref{tb:errormapping} and Figure~\ref{fig:errordistribution} in Appendix. A summary of these results is presented in Figure~\ref{fig:solution}.} Our findings reveal that: 1) LLMs like GPT-4o and Gemini 1.5 Pro can improve performance or make functional advancements in meeting scenarios by enhancing reasoning in meeting-specific context, and 2) smaller models need to improve general instruction following and reasoning abilities before addressing meeting-specific issues.

\begin{figure}[htb!]
  \centering
  \includegraphics[scale=0.33]{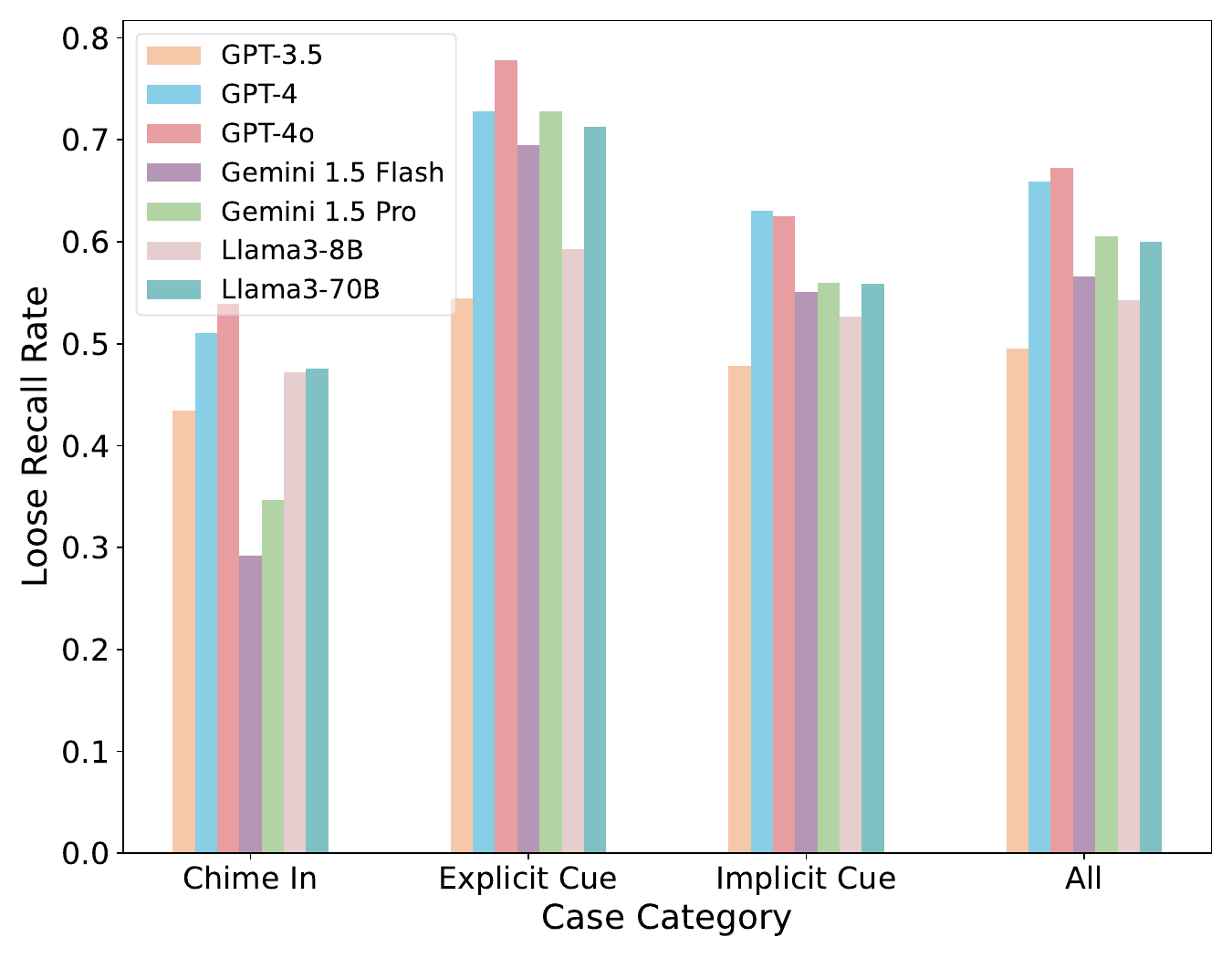}
  \caption{Loose recall rate on Matched Dataset.}
  \label{fig:loose_recall}
\end{figure}

\noindent\textbf{Recall Analysis.} The recall results for both loose and strict metrics are similar; therefore, we only present the loose recall rate for all studied LLMs on Matched Dataset in Figure~\ref{fig:loose_recall}. Detailed results, including the strict recall rate, are available in Table~\ref{tab:Recall} in the Appendix. Figure~\ref{fig:loose_recall} shows that these LLMs achieve a loose \textbf{recall rate of approximately 60\%}. This indicates that, for 60\% of test cases, the generated response contains at least one key point from the ground-truth response. Such a result is promising, as it suggests that LLM-powered meeting delegates can typically respond with reasonable content, maintaining the overall meeting flow.

Performance differences among the LLMs reveal that GPT-4o achieves the highest performance across almost all categories, followed by GPT-4. The two Gemini models exhibit similar performance, excelling in ``Explicit Cue'' but lagging in ``Chime In''. 
The Llama series models perform comparably to the Gemini models but tend to be better in ``Chime In'' scenarios.


\begin{figure}[htb!]
  \centering
  \includegraphics[scale=0.21]{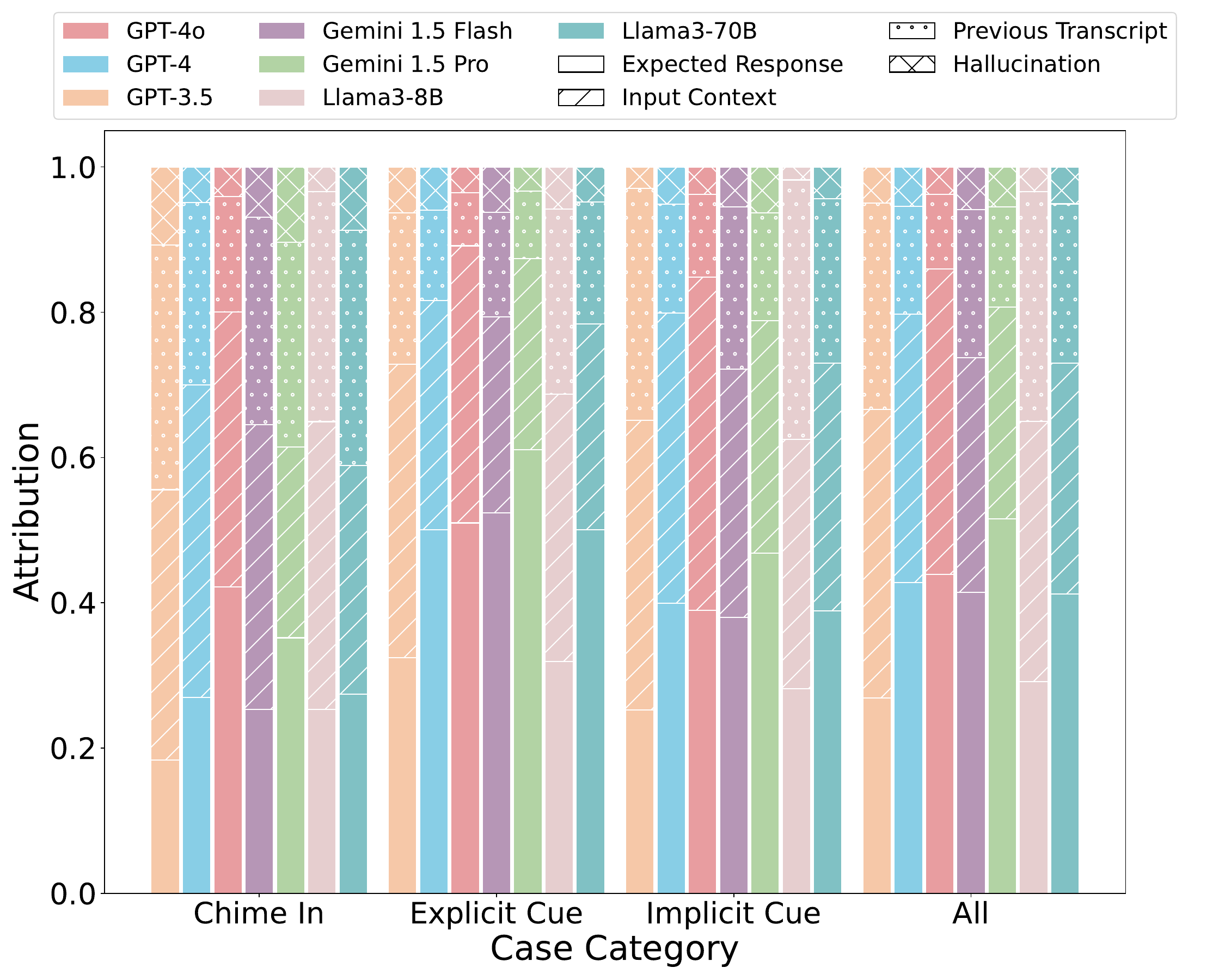}
  \caption{The attribution rate on matched dataset.}
  \label{fig:attribution}
\end{figure}

\noindent\textbf{Attribution Analysis.} For the Attribution metric, we seek a high percentage of ``Expected Response'', indicating high accuracy in responding to given cues, while minimizing other categories, particularly ``Hallucination''. As shown in Figure~\ref{fig:attribution}, most models, except GPT-3.5-Turbo and Llama3-8B, have approximately 40\% of their responses attributable to the ground-truth response, with Gemini 1.5 Pro achieving the highest performance at around 50\%. About 30\% of generated responses are attributed to other input context information not directly related to the ground-truth response, indicating room for improvement in reasoning over the provided information. The proportion attributed to the previous transcript varies significantly across models, ranging from 10\% to 30\%. Higher values suggest repetitive messages in the generated response, potentially detracting from the meeting experience due to verbosity. The portion of hallucinated texts is minimal, at only 5\% across all models, indicating that current LLMs maintain good trustworthiness in meeting engagement. 

Regarding performance differences across models, we observe that models generally considered more capable demonstrate better performance, while models like GPT-3.5-Turbo and Llama3-8B, viewed as less capable, show inferior performance. This alignment between general model performance and specific scenarios suggests that in future, more capable general LLMs will also benefit meeting delegate scenarios.


\noindent\textbf{Correlation Analysis.} We correlate the performance of the above metrics with test case metadata (\textit{i.e.}, those shown in Figure~\ref{fig:meta_data}). Figure~\ref{fig:corelation} in the Appendix presents an example result for GPT-4o. \new{The result indicates that GPT-4o maintained stable performance across different transcript lengths and complexity measures, including meeting size and input diversity. Therefore, \textit{no} significant relationships between the evaluation metrics and the metadata were observed.}


\noindent\textbf{Ablation Study.} Two scenarios are considered in our ablation studies. First, we examine the impact of erroneous transcription of participant names to phonetically similar words using the Noisy Name Dataset. We measure the response rates of all models on this dataset, observing a significant drop in performance (see Table~\ref{tab:Noisyname_Response_rate} in the Appendix). For instance, GPT-4o's response rate declines from 94.3\% in the Explicit Cue cases of the Matched Dataset to 68\% in the Noisy Name Dataset. This highlights challenges in accurately recognizing participant names. Further model fine-tuning to better handle such transcription errors may be necessary.

In our second study, we investigate how model performance is affected by the provision of context information in the input. Currently, context information is structured as pairs of <Context> and <Information>, specifying under which conditions the information in <Information> can be shared. This setup may not reflect real-world scenarios where users might not always anticipate the context for sharing specific information. To assess the impact, we remove <Context> from test cases and use <Information> and <Intents> alone as input to generate responses. We evaluate this on a subset of 121 test cases from the first 11 meetings using GPT-4o. Detailed results are provided in Table~\ref{tab:no_ctxt} in the Appendix, showing minimal performance impact across all evaluation metrics when context information is omitted.

\section{Discussion}

\begin{table*}
\centering
\small
\caption{Progression of Autonomy and Responsibility in Achieving a Fully Autonomous Meeting Delegate.}
\label{tab:phase}
\begin{tabular}{c | c  c c}
\hline
            & \textbf{Phase I: Execute} & \textbf{Phase II: Assist} & \textbf{Phase III: Delegate}\\
\hline
Data Boundary  &	User-defined boundaries	 & Privacy-protected boundaries	 &Data accessible by user\\
\hline
Share Information & \makecell{Only within\\user-defined boundaries} &	\makecell{Some reasoning over\\sensitive data} & \makecell{Autonomous based on predefined \\goals and preferences} \\
\hline
Collect Information & Explicit requests only & \makecell{Infer context beyond\\user instructions}  & \makecell{Autonomously collects and reasons \\based on meeting context}\\
\hline
Decision-Making & No decision-making & Propose and ask for approval & Full autonomous decision-making\\
\hline
\end{tabular}
\end{table*}

\new{
\subsection{Phased Deployment of Meeting Delegate}

This study primarily explores the feasibility of using LLMs to represent users by generating meaningful content in meeting scenarios. However, deploying such a meeting delegate system in real-world settings requires addressing additional critical factors, such as responsible AI practices and ethical considerations (see further discussion in the Ethics Statement section). Key challenges include implementing strong privacy safeguards, such as secure data handling, consent mechanisms, user-defined boundaries, and audit trails. A review~\cite{privacysurvey, saftysurvey} of current privacy-preserving methods for LLMs highlights the difficulty of creating a fully autonomous and unconstrained meeting delegate at present. Therefore, we propose a three-phase approach that incrementally enhances the AI's autonomy and responsibility, as detailed in Table~\ref{tab:phase}. The phases are characterized by the evolution of data boundaries and limitations on the meeting delegate’s roles in sharing information, collecting data, and making decisions.


In \textbf{Phase I (Execute)}, the delegate operates strictly within user-defined data boundaries, sharing only explicitly approved information and collecting information from other meeting participants based on direct user instructions. There is no autonomous decision-making allowed, ensuring strong user control and minimal privacy risk.
In \textbf{Phase II (Assist)}, the system can reason over sensitive data while adhering to privacy guidelines. It infers context beyond explicit instructions and can propose actions, though user approval is still required for making decisions. This phase introduces controlled autonomy with dynamic data boundary management.
In \textbf{Phase III (Delegate)}, the delegate fully autonomously collects and shares information, making real-time decisions based on user-defined goals and preferences. Privacy filters, decision-making models, and audit logs ensure transparency and accountability, with the system acting independently on behalf of the user.
This phased approach enables the delegate to transition from a controlled executor to a fully autonomous agent, balancing privacy and increasing decision-making capability while ensuring transparency and accountability.

While our ultimate goal is to achieve Phase III for significantly reducing meeting-related burdens, implementing a meeting delegate system in earlier phases may already benefit certain situations. For instance, a Phase I delegate system might be employed in daily project update scrums, where a delegate would present updates and gather progress from team members for alignment. Although one could argue that such objectives could be accomplished asynchronously via offline progress updates, deploying an early-stage system is still beneficial. It allows us to gain practical experience that will inform future advancements toward the system's full potential. Additionally, phased deployment familiarizes users with the technology, helping to identify overlooked issues and challenges.


}

\subsection{Application in Practice}
\label{sec:app_in_practice}

\new{Our current implementation of the meeting delegate system indeed corresponds to Phase I, consistent with available technologies. To assess its practical performance, the system was tested in several demo scenarios. For example, as shown in Figure~\ref{fig:demo}, we simulated a demo scenario of a daily project update scrum with three human participants and one LLM-powered delegate. All participants were aware of the delegate's presence and located in the same room. One participant acted as the moderator, while the others, including the delegate, provided project updates. Each human participant followed a script, requesting information from the delegate, which was preloaded with project-related topics via the Information Gathering module. The moderator guided the meeting, with responses cued or initiated by the participants. The demo lasted five minutes and was repeated to assess the delegate’s consistency using different LLMs.}

We evaluated three models: GPT-3.5-Turbo, GPT-4, and GPT-4o. GPT-3.5-Turbo underperformed, proving inadequate for meeting delegation tasks, even at Phase I. GPT-4 and GPT-4o generally delivered relevant responses but occasionally repeated information from earlier transcripts. Response latency was another issue, with the fastest model, GPT-4o, taking $\sim$5 seconds to respond.

To address issues of irrelevant and repetitive responses, future improvements may include utilizing advanced general LLMs or fine-tuning smaller models. Benchmark results indicate that the Llama3-8B model exhibits satisfactory base performance, and fine-tuning smaller models could potentially reduce latency. For instance, a recent implementation of Llama3-8B achieved a 500 ms latency in real-time communication~\cite{fastvoiceagent}. \new{Other improvements, such as incorporating windowed context management, advanced summarization techniques, or adopting multi-modal language models with direct speech input and output capabilities~\cite{realtimeapi}, have the potential to not only further reduce latency but also maintain or improve performance. For example, the added information from speech, such as speed and tone, could lead to enhanced system performance.}


\vspace{-5pt}
\section{Conclusion}
\vspace{-5pt}

This study introduces and evaluates an LLM-powered meeting delegate system designed to address contemporary challenges in collaborative work environments. By focusing on participant roles rather than facilitators, our prototype and comprehensive benchmark highlight the potential of LLMs to enhance meeting efficiency. Through real-world testing and rigorous assessment, we demonstrate varying performance levels among LLMs, with notable strengths and areas for improvement. Challenges include managing transcription errors and reducing irrelevant or repetitive responses. Future work will need to address these challenges and enhance the real-time responsiveness and privacy safeguards of such systems to fully realize their potential in collaborative work environments.

\newpage
\clearpage

\section*{Limitations}

We acknowledge several limitations in our study. First, the evaluation is restricted to a set of representative language models. While this provides valuable insights, future work should explore a broader range of LLMs, particularly models specifically fine-tuned for meeting-related tasks. Additionally, recent advancements such as OpenAI's Realtime API~\cite{realtimeapi}, which supports direct voice input and output, could significantly enhance the relevance of our findings in multimodal contexts.

Second, our benchmark is largely based on limited experimental conditions.
Future evaluations should incorporate more diverse and dynamic environments to provide a more comprehensive understanding of our system's capabilities.

Lastly, while our system shows promise in facilitating meeting participation, it represents an initial exploration of the possibility of using LLMs as meeting delegates. Specifically, it does not extensively address other key dimensions such as privacy, security, or user trust. 
In the following section, we share an initial discussion on responsible AI and ethics consideration to outline potential directions for further investigation.

\section*{Ethics Statement}


This paper explores the potential use of LLMs as meeting delegates, raising several ethical considerations. 
We propose a phased approach to AI autonomy, starting with limited decision-making in earlier phases and building toward greater capabilities with accountability measures. Privacy-by-design principles should be central to the system’s architecture, and educating users about the AI’s limitations will ensure responsible use. Below, we outline key ethical dimensions~\cite{ethicsdanger, ethicschatgpt, ethicshuman, ethicspersonalization}, including bias, privacy, transparency, human agency, security, and socio-economic impact, alongside suggested safeguards.

\textbf{Bias and Fairness:}
LLMs may generate biased or inappropriate content, potentially affecting fairness in meeting outcomes. This risk requires bias detection and mitigation strategies, such as training on diverse datasets, bias audits, and user feedback loops. Fine-tuning models for meeting scenarios and ongoing bias monitoring could be crucial for ensuring fairness.

\textbf{Privacy:}
Personalization is only possible by collecting user data. This applies to any technology that relies on personal information to deliver tailored benefits.
The personalization of meeting delegates relies on sensitive user data, which risks over-sharing or misusing private information. To address this, we advocate for privacy-enhancing technologies like encryption and differential privacy, as well as user-defined data boundaries. Real-time voice capabilities also heighten the risk of identity misuse, necessitating strict privacy controls to ensure compliance with data protection standards.

\textbf{Transparency:}
Transparency is essential for responsible deployment. All participants must be informed when an AI is acting as a delegate. Clearly stating the AI's capabilities and limitations helps manage expectations, and audit logs should be available for users to track AI actions and decisions during meetings.

\textbf{Human Agency:}
LLM-based delegates should support, not replace, human decision-making. In the early phases, the AI assists users without autonomy, and even in later phase like Phase III, human oversight must remain integral. Human-in-the-loop HITL systems are crucial for maintaining control and ensuring users can intervene as needed.

\textbf{Security and Fraud Risks:}
Unauthorized access to a meeting delegate could lead to fraud or impersonation. Security measures like multi-factor authentication, identity verification, and anomaly detection are essential. Federated learning could further protect sensitive data by minimizing centralized storage risks.

\textbf{Ethical Governance and Mitigation:}
Ethical governance frameworks, including guidelines, audits, and interdisciplinary collaboration, must guide the system’s development. User consent should be obtained at key stages, and continuous monitoring is essential to identify and address unintended consequences.

\textbf{Socio-Economic Impact:}
Automating meeting participation could lead to job displacement in roles that rely on meeting facilitation. While this risk is limited by current technology, future developments may amplify these concerns. It's essential to focus on augmenting human labor rather than replacing.


\bibliography{custom}

\appendix

\section{Dataset Construction}

\begin{figure*}[htb!]
  \centering
  \includegraphics[width=\textwidth]{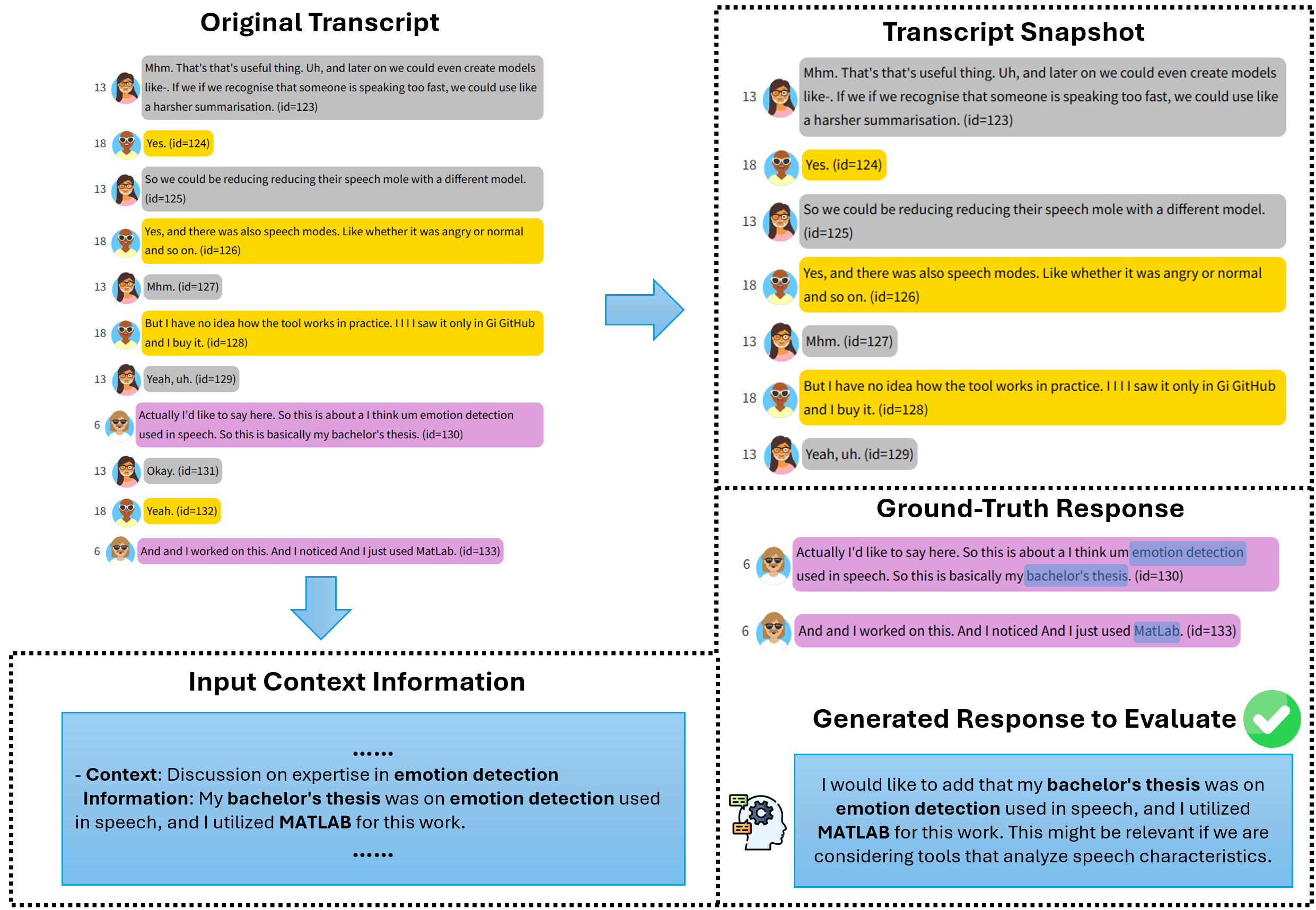}
  \caption{Example of evaluation dataset construction. Participants are represented by different ID numbers and icons. Colored boxes indicate utterances from different participants. The process includes extracting Input Context Information, creating a Transcript Snapshot, and generating a response with the LLM-powered meeting delegate. The Generated Response is evaluated by comparison with the Ground-Truth Response.}
  \label{fig:dataset_construction}
\end{figure*}

An example of evaluation dataset construction is shown in Figure~\ref{fig:dataset_construction}. In the meeting transcript, participants are represented by different ID numbers and icons. Each utterance is displayed in colored boxes, with each color representing a different participant.
In this example, we construct a test case with Participant 6 as the principal. Based on Participant 6's utterances in the Original Transcript, we extract one piece of Input Context Information: when the meeting discusses expertise in emotion detection, Participant 6 intends to mention related experience from bachelor thesis.
The Transcript Snapshot and Ground-Truth Response are extracted from the Original Transcript using GPT-4. During the response generation stage with the meeting delegate, the Transcript Snapshot is provided to the LLMs to produce a response. This generated response is subsequently assessed by comparing it to the Ground-Truth Response.

We plan to release our constructed benchmark dataset with the paper.

\section{Additional Experimental Results}
\label{sec:appendix_exp_results}

In this section, we provide detailed tables and additional plots for the experimental results discussed in Section~\ref{sec:experiment}.

\noindent\textbf{Response Rate Analysis.} 
Tables~\ref{tab:Response_rate} and~\ref{tab:Mismatch_Response_rate} present the Response Rate and Silence Rate of LLMs evaluated using the Matched and Mismatched Datasets, respectively.
Additionally, in Tables~\ref{tab:Response_rate_inter} and~\ref{tab:Mismatch_Response_rate_inter}, we further evaluate the Response Rate and Silence Rate using the intersection subdataset of all models, given that Llama models and GPT-3.5 have smaller context windows. The findings from these experimental results remain consistent.

\begin{figure*}[t]
\centering
\subfigure[Chine In (Matched Dataset)]{
	\includegraphics[width=.6\linewidth]{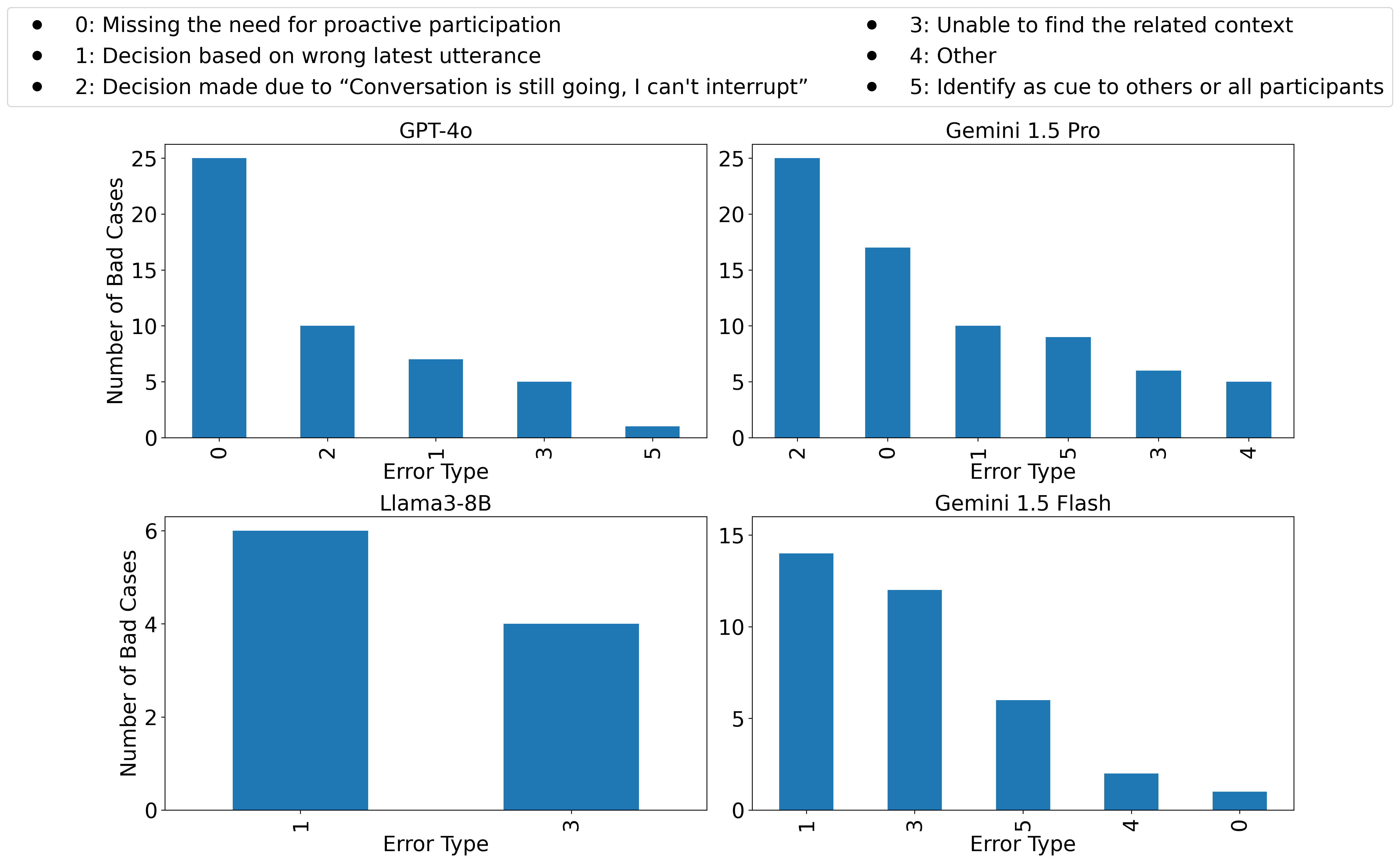}
    \label{subfig:lifetime}
}
\subfigure[Explicit Cue (Matched Dataset)]{
	\includegraphics[width=.65\linewidth]{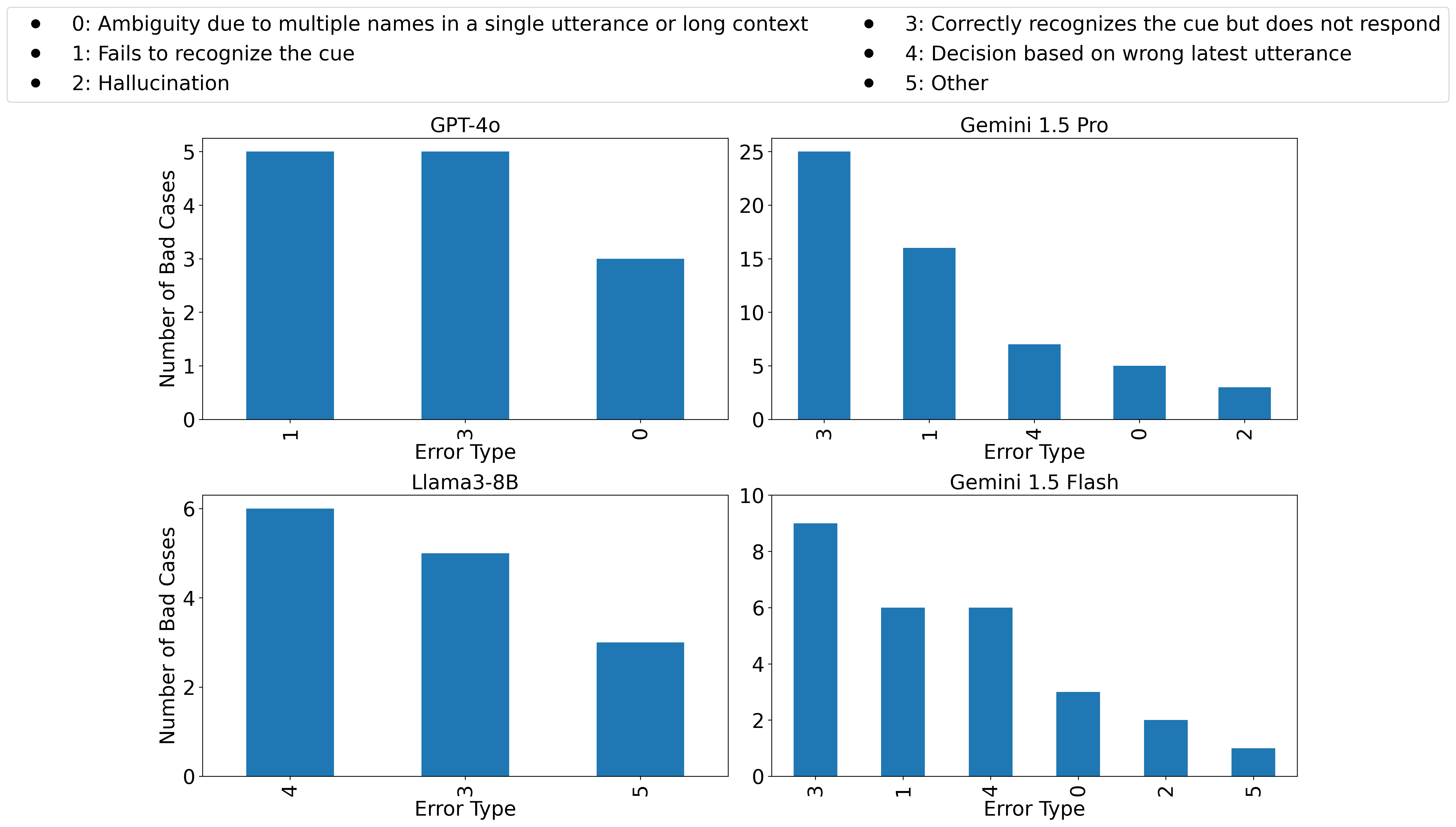}
    \label{subfig:vm_cnt_variation}
}
\subfigure[Mismatched Dataset]{
	\includegraphics[width=.55\linewidth]{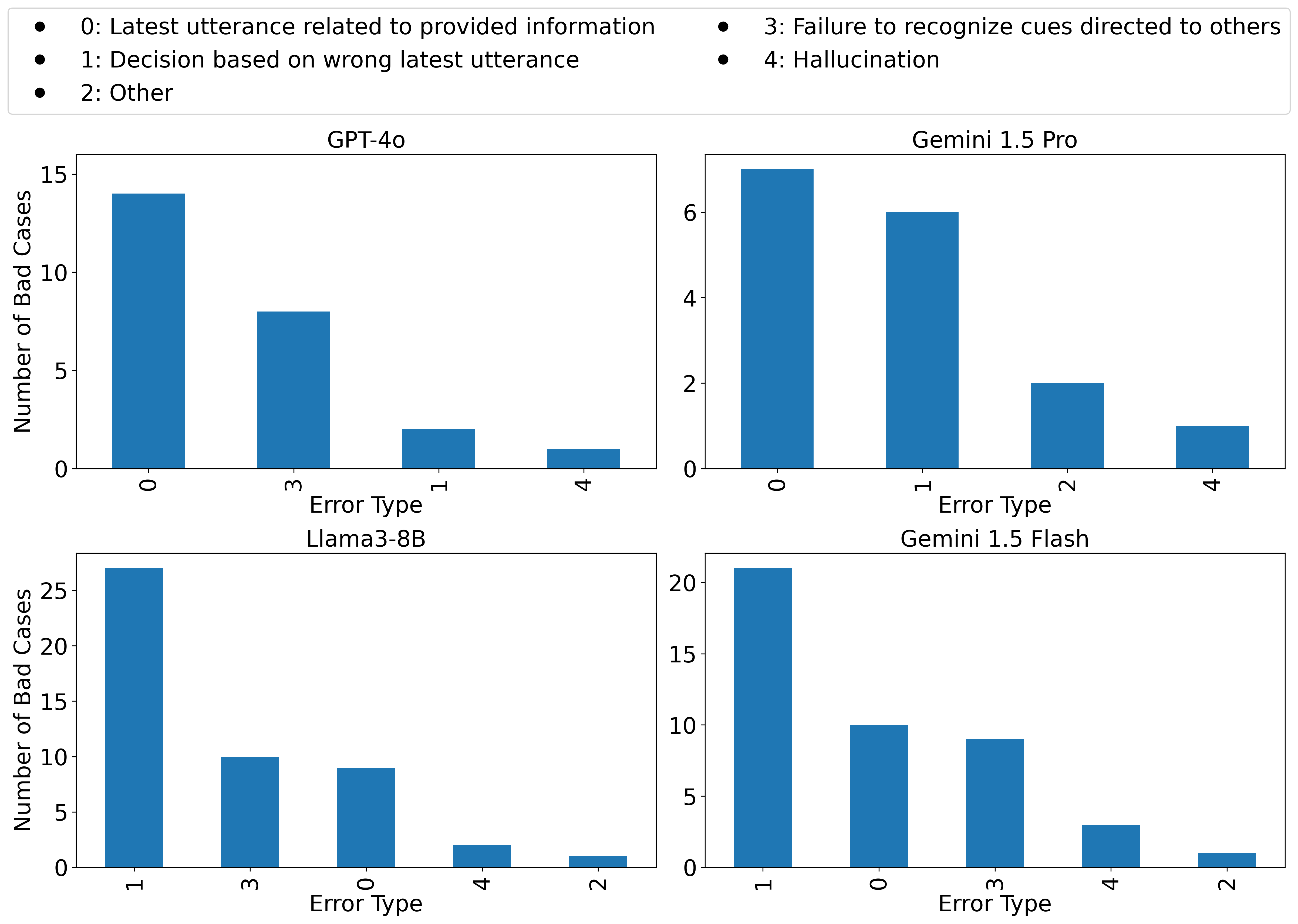}
    \label{subfig:vm_creation_variation}
}

\caption{
(a) Error Types Distribution for Response Rate Failure Cases Study in Chine In Matched Dataset.
(b) Error Types Distribution for Response Rate Failure Cases Study in Explicit Cue Matched Dataset.
(c) Error Types Distribution for Response Rate Failure Cases Study in Mismatched Dataset.
}

\label{fig:errordistribution}
\end{figure*}

\noindent\textbf{Response Rate Failure Cases Study.}
The error types distribution for response rate failure cases study in Matched and Mismatched datasets are presented in Figure~\ref{fig:errordistribution}. The mappings between error types and improvement solution direction are summarized in Table~\ref{tb:errormapping}.

\noindent\textbf{Recall Analysis.} 
The loose recall rate and strict recall rate for the Matched Dataset are shown in Table \ref{tab:Recall}. We further evaluate the recall rates using the intersection subdataset of all models, with results presented in Table~\ref{tab:Recall_inter}. Although the absolute values of recall rates for all models are higher, the performance differences among the models are similar. Note that we do not include Llama3-8B and Llama3-70B here in the intersection study to avoid too few samples. The findings from these experimental results remain consistent.

\noindent\textbf{Attribution Analysis.} 
The attribution metrics for LLMs are included in Table~\ref{tab:Attribution}. We also evaluate the attribution metrics using the intersection subdataset. Note that we do not include Llama3-8B and Llama3-70B here in the intersection study to avoid too few samples. The findings from these experimental results remain consistent.

\begin{figure*}[htb!]
  \centering
  \includegraphics[scale=0.26]{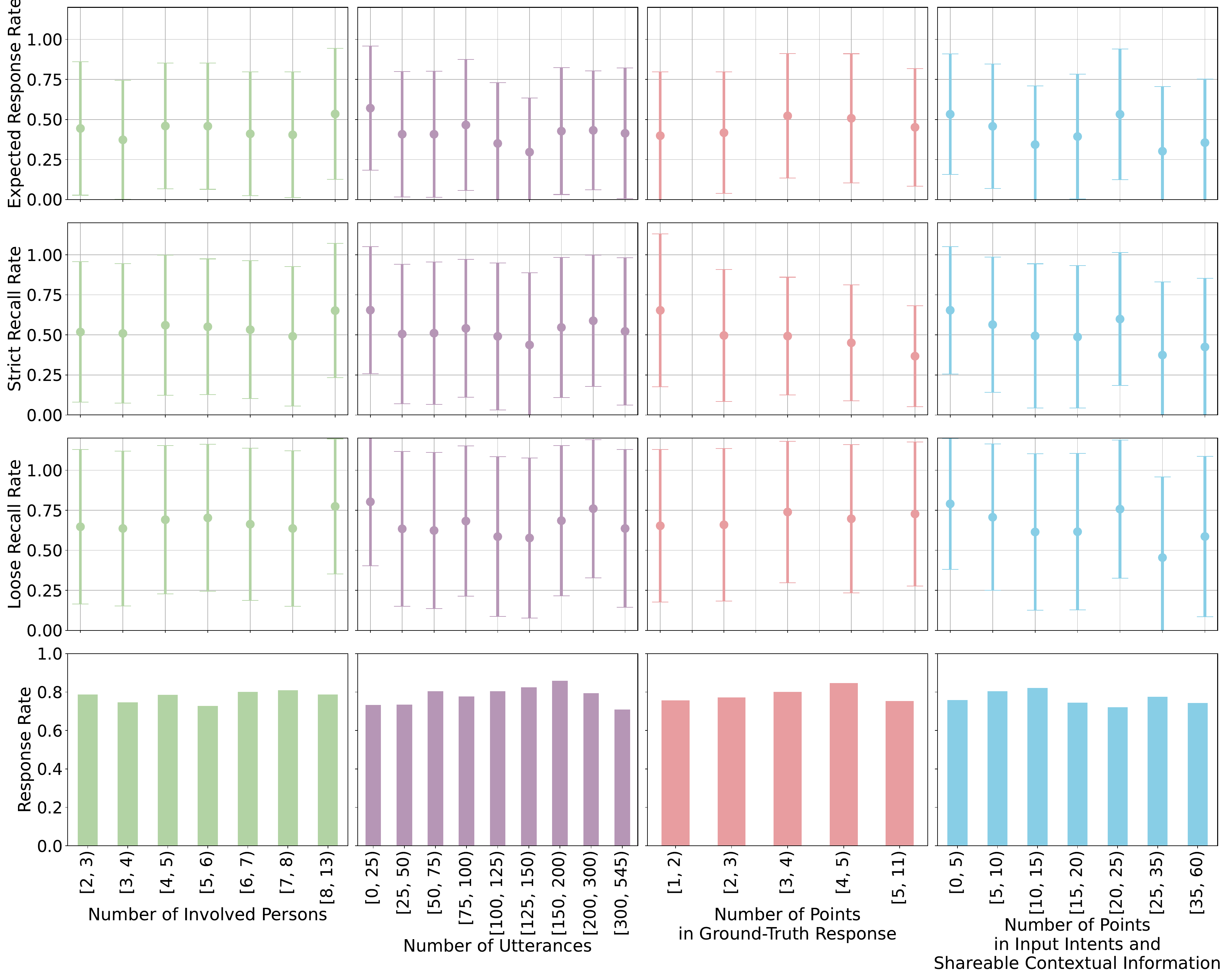}
  \caption{The correlation between the performance metrics and test case metadata for GPT-4o.}
  \label{fig:corelation}
\end{figure*}

\noindent\textbf{Correlation Study.} 
The correlation of response rate and recall metrics with test case metadata is shown in Figure~\ref{fig:corelation}.
No significant correlations is found between these metrics and the metadata.

\noindent\textbf{Ablation Study.} 
The response rates of LLMs for the Noisy Name Dataset are presented in Table \ref{tab:Noisyname_Response_rate}, with the response rates from Explicit Cue in Matched Dataset are also shown for reference. A significant drop in
performance is observed for all models, except for GPT-3.5 where responses rates are already low. This further
highlights challenges in accurately recognizing participant names. Further model fine-tuning to better handle such transcription errors may be necessary.
For the No-<Context> study, all evaluation metrics for GPT-4o in No-<Context> Scenario are shown in Table \ref{tab:no_ctxt}, showing minimal performance impact across all evaluation metrics when context information is omitted.

\begin{table*}
\centering
\small
\caption{Response Rate for Matched Dateset.}
\label{tab:Response_rate}
\begin{tabular}{cccccccc}
\toprule
Type            & GPT-3.5 & GPT-4 & GPT-4o & Gemini 1.5 Flash & Gemini 1.5 Pro & Llama3-8B & Llama3-70B\\
\midrule
Chime In   & 39.3\%   & 37.9\% & 61.3\%  & 71.8\%   & 41.9\% & 84.1\% & \textbf{93.8\%}   \\
Explicit Cue & 53.2\%  & 86.7\% & 94.3\% & 89.7\%   & 78.3\% & 91.2\% & \textbf{99.4\%}   \\
Implicit Cue & 52.2\%  & 67.2\% & 71.9\% & 83.6\%   & 55.9\% & 90.0\% & \textbf{94.8\%}    \\
All & 50.6\%  & 68.9\% & 77.3\% & 83.8\% & 60.8\% & 89.6\% & \textbf{96.2\%}      \\
\bottomrule
\end{tabular}
\end{table*}

\begin{table*}
\centering
\small
\caption{Response Rate for Intersection Subset of Matched Dateset.}
\label{tab:Response_rate_inter}
\begin{tabular}{cccccccc}
\toprule
Type            & GPT-3.5 & GPT-4 & GPT-4o & Gemini 1.5 Flash & Gemini 1.5 Pro & Llama3-8B & Llama3-70B\\
\midrule
Chime In   & 35.2\%   & 42.3\% & 57.7\%  & 66.2\%   & 43.7\% & 81.7\% & \textbf{95.8\%}   \\
Explicit Cue & 58.6\%  &92.0\% & 92.0\% & 87.7\%   & 76.5\% & 89.5\% & \textbf{98.1\%}   \\
Implicit Cue & 54.3\%  & 65.8\% & 68.3\% & 81.9\%   & 53.5\% & 89.7\% & \textbf{94.7\%}    \\
All & 52.9\%  & 71.2\% & 74.8\% & 81.5\% & 59.9\% & 88.4\% & \textbf{96.0\%}      \\
\bottomrule
\end{tabular}
\end{table*}

\begin{table*}
\centering
\small
\caption{Silence Rate for Mismatched Dataset.}
\label{tab:Mismatch_Response_rate}
\begin{tabular}{cccccccc}
\toprule
Type            & GPT-3.5 & GPT-4 & GPT-4o & Gemini 1.5 Flash & Gemini 1.5 Pro & Llama3-8B & Llama3-70B\\
\midrule
Explicit Cue & 75.0\%  & 84.6\% & 82.8\% & 65.0\%   & \textbf{88.1\%} & 36.0\% & 41.6\%   \\
Implicit Cue & 70.4\%  & \textbf{79.5\%} & 67.9\% & 52.0\%   & 77.1\% & 35.3\% & 33.3\%    \\
All & 72.4\%  & 81.6\% & 73.6\% & 57.5\% & \textbf{81.7\%} & 35.6\% & 37.0\%      \\
\bottomrule
\end{tabular}
\end{table*}

\begin{table*}
\centering
\small
\caption{Silence Rate for Intersection Subset of Mismatched Dataset.}
\label{tab:Mismatch_Response_rate_inter}
\begin{tabular}{cccccccc}
\toprule
Type            & GPT-3.5 & GPT-4 & GPT-4o & Gemini 1.5 Flash & Gemini 1.5 Pro & Llama3-8B & Llama3-70B\\
\midrule
Explicit Cue & 79.5\%  & 84.9\% & \textbf{90.4\%} & 76.7\%   & \textbf{90.4\%} & 37.0\% & 44.6\%   \\
Implicit Cue & 69.5\%  & \textbf{81.7\%} & 74.4\% & 58.5\%   & \textbf{81.7\%} & 35.4\% & 31.9\%    \\
All & 74.2\%  & 83.2\% & 81.9\% & 67.1\% & \textbf{85.8\%} & 36.1\% & 38.7\%      \\
\bottomrule
\end{tabular}
\end{table*}

\begin{table*}
\centering
\small
\caption{Recall Rate for Matched Dataset.}
\label{tab:Recall}
\begin{tabular}{lcccccccc}
\toprule
\multirow{2}{*}{Model} & \multicolumn{2}{c}{Chime In} & \multicolumn{2}{c}{Explicit Cue} & \multicolumn{2}{c}{Implicit Cue} & \multicolumn{2}{c}{All} \\
\cmidrule(lr){2-3} \cmidrule(lr){4-5} \cmidrule(lr){6-7} \cmidrule(lr){8-9}
& Loose & Strict & Loose & Strict & Loose & Strict & Loose & Strict \\
\midrule
GPT-3.5 & 43.5\% & 29.5\% & 54.5\% & 42.5\%  & 47.8\% & 37.0\% & 49.5\% & 38.0\%\\
GPT-4 & 51.1\% & 39.9\% & 72.8\% & 60.7\%  & \textbf{63.0\%} & \textbf{49.6\%} & 65.9\% & 53.1\%\\
GPT-4o & \textbf{53.9\%} & \textbf{47.0\%} & \textbf{77.8\%} & \textbf{64.2\%} & 62.5\% & 47.9\% & \textbf{67.3\%} & \textbf{53.9\%} \\
Gemini 1.5 Flash & 29.2\% & 22.5\% & 69.5\% & 56.5\%  & 55.0\% & 40.2\% & 56.6\% & 43.4\%\\
Gemini 1.5 Pro & 34.6\% & 28.8\% & 72.8\% & 59.9\%  & 56.0\% & 43.5\% & 60.5\% & 48.6\%\\
Llama3-8B & 46.7\% & 35.5\% & 59.6\% & 48.7\%  & 52.7\% & 40.5\% & 54.2\% & 42.6\%\\
Llama3-70B & 45.8\% & 34.7\% & 69.6\% & 59.4\%  & 55.9\% & 44.0\% & 59.1\% & 47.9\%\\
\bottomrule
\end{tabular}
\end{table*}

\begin{table*}
\centering
\small
\caption{Recall Rate for Intersection Subset of Matched Dataset. Note that due to limited statistics for intersecting Llama results, Llama results are not included. The total number of cases in the considered Intersection Subset is 196.}
\label{tab:Recall_inter}
\begin{tabular}{lcccccccc}
\toprule
\multirow{2}{*}{Model} & \multicolumn{2}{c}{Chime In} & \multicolumn{2}{c}{Explicit Cue} & \multicolumn{2}{c}{Implicit Cue} & \multicolumn{2}{c}{All} \\
\cmidrule(lr){2-3} \cmidrule(lr){4-5} \cmidrule(lr){6-7} \cmidrule(lr){8-9}
& Loose & Strict & Loose & Strict & Loose & Strict & Loose & Strict \\
\midrule
GPT-3.5 & 55.6\% & 47.2\% & 58.4\% & 46.9\%  & 56.1\% & 45.2\% & 57.1\% & 46.0\%\\
GPT-4 & \textbf{77.8\%} & \textbf{52.8\%} & 79.8\% & 66.7\%  & 70.4\% & 55.8\% & 75.0\% & 60.6\%\\
GPT-4o & 66.7\% & \textbf{52.8\%} & \textbf{85.4\%} & \textbf{70.6\%} & \textbf{79.6\%} & \textbf{59.8\%} & \textbf{81.6\%} & \textbf{64.4\%} \\
Gemini 1.5 Flash & 44.4\% & 32.2\% & 79.8\% & 64.6\%  & 67.3\% & 49.3\% & 71.9\% & 55.4\%\\
Gemini 1.5 Pro & 22.2\% & 19.4\% & 77.5\% & 62.6\%  & 60.2\% & 46.2\% & 66.3\% & 52.4\%\\
\bottomrule
\end{tabular}
\end{table*}

\begin{table*}[ht]
\centering
\small
\caption{Attribution Analysis results for Matched Dataset. For the Expected Response metric, higher values are better, while for the Previous Transcript and Hallucination metrics, lower values are preferable.}
\label{tab:Attribution}
\begin{tabular}{cccccccc}
\toprule
Metric & GPT-3.5 & GPT-4 & GPT-4o & Gemini 1.5 Flash & Gemini 1.5 Pro & Llama3-8B & Llama3-70B \\
\midrule
\multicolumn{8}{c}{Chime In} \\
\cmidrule(lr){1-8}
Expected Response & 18.4\% & 27.0\% & \textbf{42.2\%} & 25.3\% & 35.2\% & 25.4\% & 27.4\%\\
Input Context Info & 37.2\% & 43.0\% & 37.9\% & 39.2\% & 26.3\% & 39.6\% & 31.4\%\\
Previous Transcript & 33.7\% & 25.1\% & \textbf{15.9\%} & 28.6\% & 28.1\% & 31.6\% & 32.4\%\\
Hallucination & 10.8\% & 4.93\% & 4.05\% & 6.93\% & 10.4\% & \textbf{3.43\%} & 8.75\%\\
\midrule
\multicolumn{8}{c}{Explicit Cue} \\
\cmidrule(lr){1-8}
Expected Response & 32.5\% & 50.1\% & 51.0\% & 52.4\% & \textbf{61.1\%} & 31.9\% & 50.1\%\\
Input Context Info & 40.4\% & 31.6\% & 38.1\% & 27.0\% & 26.3\% & 36.8\% & 28.3\%\\
Previous Transcript & 20.8\% & 12.4\% & \textbf{7.28\%} & 14.4\% & 9.25\% & 25.4\% & 16.8\%\\
Hallucination & 6.43\% & 5.98\% & 3.58\% & 6.24\% & \textbf{3.35\%} & 5.82\% & 4.81\%\\
\midrule
\multicolumn{8}{c}{Implicit Cue} \\
\cmidrule(lr){1-8}
Expected Response & 25.2\% & 39.9\% & 38.9\% & 38.0\% & \textbf{46.8\%} & 28.2\% & 38.9\%\\
Input Context Info & 39.9\% & 39.9\% & 45.9\% & 34.2\% & 32.0\% & 34.3\% & 34.1\%\\
Previous Transcript & 31.9\% & 15.0\% & \textbf{11.3\%} & 22.4\% & 14.8\% & 35.7\% & 22.6\%\\
Hallucination & 2.96\% & 5.12\% & 3.8\% & 5.48\% & 6.32\% & \textbf{1.80\%} & 4.38\%\\
\midrule
\multicolumn{8}{c}{All} \\
\cmidrule(lr){1-8}
Expected Response & 26.9\% & 42.8\% & 43.9\% & 41.5\% & \textbf{51.6\%} & 29.1\% & 41.2\%\\
Input Context Info & 39.8\% & 36.9\% & 42.0\% & 32.3\% & 29.2\% & 35.8\% & 31.8\%\\
Previous Transcript & 28.4\% & 14.8\% & \textbf{10.3\%} & 20.3\% & 13.8\% & 31.6\% & 21.9\%\\
Hallucination & 4.95\% & 5.44\% & 3.74\% & 5.91\% & 5.48\% & \textbf{3.39\%} & 5.09\%\\
\bottomrule
\end{tabular}
\end{table*}

\begin{table*}[ht]
\centering
\small
\caption{Attribution Analysis results for Intersection Subset of Matched Dataset. For the Expected Response metric, higher values are better, while for the Previous Transcript and Hallucination metrics, lower values are preferable. Note that due to limited statistics for the intersecting Llama results, Llama results are not included. The total number of cases in the considered Intersection Subset is 196.}
\label{tab:Attribution_inter}
\begin{tabular}{cccccc}
\toprule
Metric & GPT-3.5 & GPT-4 & GPT-4o & Gemini 1.5 Flash & Gemini 1.5 Pro \\
\midrule
\multicolumn{6}{c}{Chime In} \\
\cmidrule(lr){1-6}
Expected Response & 22.0\% & 30.9\% & \textbf{35.8} & 29.2\% & 22.2\% \\
Input Context Info & 55.7\% & 58.1\% & 64.2\% & 45.8\% & 22.2\% \\
Previous Transcript & 11.1\% & 5.0\% & \textbf{0.0\%} & 12.5\% & 44.4\% \\
Hallucination & 11.1\% & 5.9\% & \textbf{0.0\%} & 12.5\% & 11.1\% \\
\midrule
\multicolumn{6}{c}{Explicit Cue} \\
\cmidrule(lr){1-6}
Expected Response & 37.4\% & 59.1\% & 56.1\% & 59.9\% & \textbf{66.9\%} \\
Input Context Info & 37.9\% & 27.7\% & 36.4\% & 23.3\% & 19.5\% \\
Previous Transcript & 19.6\% & 10.1\% & \textbf{3.3\%} & 11.7\% & 12.5\% \\
Hallucination & 5.1\% & 5.98\% & 3.1\% & 5.1\% & \textbf{1.2\%} \\
\midrule
\multicolumn{6}{c}{Implicit Cue} \\
\cmidrule(lr){1-6}
Expected Response & 30.6\% & 47.3\% & 49.9\% & 49.4\% & \textbf{51.3\%} \\
Input Context Info & 42.6\% & 36.4\% & 38.6\% & 31.3\% & 29.4\% \\
Previous Transcript & 23.5\% & 12.2\% & \textbf{7.0\%} & 17.6\% & 12.1\% \\
Hallucination & 3.3\% & 4.0\% & 4.5\% & \textbf{1.7\%} & 7.1\% \\
\midrule
\multicolumn{6}{c}{All} \\
\cmidrule(lr){1-6}
Expected Response & 33.3\% & 51.8\% & 52.1\% & 53.4\% & \textbf{57.0\%} \\
Input Context Info & 41.1\% & 33.5\% & 38.8\% & 28.2\% & 24.5\% \\
Previous Transcript & 21.2\% & 10.9\% & \textbf{5.0\%} & 14.7\% & 13.8\% \\
Hallucination & 4.5\% & \textbf{3.7\%} & 4.1\% & \textbf{3.7\%} & 4.6\% \\
\bottomrule
\end{tabular}
\end{table*}

\begin{table*}
\centering
\small
\caption{Response rate for Noisy Name Dataset.}
\label{tab:Noisyname_Response_rate}
\begin{tabular}{ccccccccc}
\toprule
Type & Dataset            & GPT-3.5 & GPT-4 & GPT-4o & Gemini 1.5 Flash & Gemini 1.5 Pro & Llama3-8B & Llama3-70B\\
\midrule
Explicit Cue & Matched & 53.2\%  & 86.7\% & 94.3\% & 89.7\%   & 78.3\% & 91.2\% & \textbf{99.4\%}   \\
Explicit Cue & Noisy Name & 52.5\%  & 53.3\% & 68.0\% & 60.7\%  & 59.8\% & 79.4\% & \textbf{87.0\%}\\ 
\bottomrule
\end{tabular}
\end{table*}

\begin{table*}
\centering
\small
\caption{All Evaluation Metrics for GPT-4o in No-<Context> Scenario.}
\label{tab:no_ctxt}
\begin{tabular}{lcccc}
\toprule
Metric & Chime In & Explicit Cue & Implicit Cue & All \\
\midrule
Response Rate       & 59.1\% & 90.4\% & 78.7\% & 80.2\% \\
Loose Recall        & 46.2\% & 82.6\% & 75.0\% & 74.7\% \\
Strict Recall       & 37.7\% & 65.0\% & 50.2\% & 55.8\% \\
Expected Response   & 21.0\% & 44.1\% & 44.7\% & 41.1\% \\
Input Context Info  & 57.2\% & 36.0\% & 31.9\% & 37.3\% \\
Previous Transcript & 14.1\% & 14.4\% & 14.0\% & 14.2\% \\
Hallucination       & 7.7\%  & 5.4\%  & 9.4\%  & 7.3\%  \\
\bottomrule
\end{tabular}
\end{table*}

\begin{table*}[t]
  \centering
  \scriptsize
  \caption{Mapping between Error Types and Solution Direction for Response Rate Failure Cases Study.}
  \begin{tabular}{c | c | c | c }
  \hline
  \hline
  Dataset & Scenarios & Error Type & Solution Direction \\
  \hline
  \multirow{12}{*}{Matched} & \multirow{6}{*}{Chime In} & Decision based on wrong latest utterance & Improved Instruction Following \\
  & & Identify as cue to others or all participants & Enhanced Reasoning in Meeting Scenario \\
  & & Missing the need for proactive participation & Enhanced Reasoning in Meeting Scenario\\
  & & Decision made due to “Conversation is still going, I can't interrupt” &Enhanced Reasoning in Meeting Scenario\\
  & & Unable to find the related context & Enhanced General Reasoning\\
  & & Other & N/A\\
  \cline{2-4}
  & \multirow{6}{*}{Explicit Cue} & Decision based on wrong latest utterance & Improved Instruction Following\\
  & & Correctly recognizes the cue but does not respond& Enhanced Reasoning in Meeting Scenario\\
  & & Ambiguity due to multiple names in a single utterance or long context &Enhanced Reasoning in Meeting Scenario\\
  & & Fails to recognize the cue  & Enhanced General Reasoning\\
  & & Hallucination & Enhanced General Reasoning\\
  & & Other & N/A\\
  \hline
  \multirow{5}{*}{Mismatched} & \multirow{5}{*}{Mismatched} & Decision based on wrong latest utterance & Improved Instruction Following\\
  & & Latest utterance related to provided information & Enhanced Reasoning in Meeting Scenario\\
  & & Failure to recognize cues directed to others & Enhanced Reasoning\\
  & & Hallucination & Enhanced General Reasoning\\
  & & Other & N/A\\

  \hline
  \hline
  
  \end{tabular}

  \label{tb:errormapping}
\end{table*}

\section{Model Specifications}
In Table~\ref{tab:model_use}, we list all LLMs utilized in this paper, along with their detailed model version and usage scenarios.

\begin{table*}
\centering
\scriptsize
\caption{Details of Model Use Scenarios and Model Version.}
\label{tab:model_use}
\begin{tabular}{l | c | c}
\hline
\hline
Model Name & Model Use Scenarios & Model Version \\
\hline
GPT-3.5       & \makecell{Generate Response (Table \ref{tab:Response_rate} \& Table \ref{tab:Response_rate_inter} \& Table \ref{tab:Mismatch_Response_rate} \\ \& Table \ref{tab:Mismatch_Response_rate_inter}, Prompt in Table \ref{tab:prompt_generate_response})} & gpt-3.5-turbo-1106 with 16k context window\\
\hline
\multirow{5}{*}{GPT-4} & \makecell{Generate Response (Table \ref{tab:Response_rate} \& Table \ref{tab:Response_rate_inter} \& Table \ref{tab:Mismatch_Response_rate} \\ \& Table \ref{tab:Mismatch_Response_rate_inter}, Prompt in Table \ref{tab:prompt_generate_response})}  & gpt-4-turbo-20240409  with 128k context window\\
\cline{2-3}
& Evaluation (Table \ref{tab:Recall} \& Table \ref{tab:Recall_inter}, Prompt in Table \ref{tab:prompt_evaluation}) & gpt-4-1106-preview with 128k context window \\
\cline{2-3}
& Attribution (Table \ref{tab:Attribution} \& Table \ref{tab:Attribution_inter}, Prompt in Table \ref{tab:prompt_attribution}) & \multirow{3}{*}{gpt-4-turbo-20240409  with 128k context window}\\
& Extract context information (Figure \ref{fig:dataset_construction}, Prompt in Table \ref{tab:prompt_context_extraction}) &  \\
& Extract test cases (Figure \ref{fig:dataset_construction}, Prompt in Table \ref{tab:prompt_case_extraction}) &  \\
\hline
GPT-4o        & \makecell{Generate Response (Table \ref{tab:Response_rate} \& Table \ref{tab:Response_rate_inter} \& Table \ref{tab:Mismatch_Response_rate} \\ \& Table \ref{tab:Mismatch_Response_rate_inter}, Prompt in Table \ref{tab:prompt_generate_response})} & gpt-4o-20240513-preview with 128k context window\\
\hline
\hline
\end{tabular}
\end{table*}

\section{Prompts}
\label{sec:appendix_prompts}
We include all prompts used in the paper. Table~\ref{tab:prompt_generate_response} provides the prompt for generating the response in the Meeting Engagement module. The prompts used for evaluating and attributing the generated response are given in Tables~\ref{tab:prompt_evaluation} and \ref{tab:prompt_attribution}, respectively. Lastly, the prompts for extracting context information and extracting test cases from meeting transcripts are given in Table~\ref{tab:prompt_context_extraction} and Table~\ref{tab:prompt_case_extraction}, respectively.

\begin{table*}[htbp]
    \centering
    \footnotesize
    \begin{tabular}{p{15cm}}
    \toprule
    - You are a Meeting Delegate Agent that attends meetings on behalf of <Person Name>.\\
    - You are provided with the intent of participating in the meeting, specified as <Intents>.\\
    - You are provided with the background information that <Person Name> knows, specified as <Background>.\\
    - You are provided with the full list of attendees <Attendees> to help identify if someone cues you.\\
    - You are provided with the ongoing meeting transcript <Meeting Transcript> to determine if there is a need to respond.\\
    - Your task is to assess the content of the ongoing meeting transcript <Meeting Transcript> and determine whether you are can speak and what to say.\\
    - You are encouraged to respond and ask questions, give comments, or share information without interrupting others in the meeting.\\
    \\
    \#\# About the <Person Name>\\
    - <Person Name> is the name of the person you represent in the meeting.\\
    - People in the <Attendees> list may cue you by using <Person Name> exactly or parts of the name (e.g., first name, initials).\\
    \\
    \#\# About the <Attendees>\\
    - <Attendees> is a list of names of the people attending the meeting.\\
    - Each name in the list is a full name or a nickname.\\
    \\
    \#\# About the <Meeting Transcript>\\
    - <Meeting Transcript> is a series of utterances spoken by the meeting participants.\\
    - Each utterance is formatted as "Name: Content", where 'Name' is the speaker's name and 'Content' is their spoken text.\\
    - The utterances are in chronological order and the latest utterance is at the bottom of the transcript.\\
    - The utterances may contain typos and grammatical errors.\\
    \\
    \#\# About the <Intents>\\
    - <Intents> consists of the questions or topics that <Person Name> aims to discuss during the meeting.\\
    - You can ask the questions or motivate the discussion of the topics in the <Intents> at the appropriate time without interrupting others.\\
    \\
    \#\# About the <Background>\\
    - <Background> consists of the background information that <Person Name> knows before the meeting.\\
    - <Background> is a list of "Context" and "Information" pairs. You can share the "Information" in the "Context" at the appropriate time without interrupting others.\\
    \\
    \#\# Guidelines to judge whether you can speak and decide what to say\\
    - Read the <Meeting Transcript> to understand the context of the meeting.\\
    - Focus on the latest several utterances in the <Meeting Transcript> to understand the current discussion.\\
    - Remember that you are a delegate attending the meeting on behalf of <Person Name>.\\
    - You should judge whether you can speak first, then decide what to say, if you can speak.\\
    - Judge whether you can speak according to the following instructions:\\
    ~~~~- Figure out what the latest utterance (at the bottom of the <Meeting Transcript>) is about and pay attention to who is being addressed.\\
    ~~~~- If the latest utterance is a straightforward question or request or instructions to other participants, you MUST NOT speak to avoid interrupting others, even if the conversation is related to the <Intents> or <Background>.\\
    ~~~~- If the latest utterance is for the <Person Name>, you should respond to it.\\
    ~~~~- If you can speak, consider the following guidelines:\\
    ~~~~- Your speech content should be directly relevant to the current discussion.\\
    ~~~~- You can reference the <Intents> and <Background> to organize your speech.\\
    ~~~~- You should be polite and natural in your speech.\\
    ~~~~- You MUST NOT make up facts.\\
    ~~~~- You MUST NOT repeat what <Person Name> has said in the <Meeting Transcript>.\\
    ~~~~- Chit chat is a natural part of conversation. You can engage in chit chat with other attendees if it is appropriate or relevant to the meeting context. For example, you can say good morning, Thank you, Yeah, I agree.\\
    ~~~~- Before speaking, you should think twice to ensure that you are not interrupting others and your speech is relevant to the current discussion.\\
    \\
    \#\# Notes on judging whether someone is cued\\
    - The name may be transcribed as similar-sounding words by the speech recognition system. Especially, the pronunciation of Chinese names may be recognized as similar-sounding English words, for exmaple, "Si Li" may be transcribed as "Celine" or "silence".\\
    - When encountering words that seem out of place, it is likely due to errors in speech recognition. Examine the list of attendees to determine if the pronunciation of these words are similar to any English or Chinese names listed.\\
    - You should consider the context of the meeting and the names of the attendees to determine if you or someone are cued.\\
    \\

    \textbf{CONTINUE ON THE NEXT PAGE} \\
    
    \bottomrule
    \end{tabular}
    \caption{Prompt used for generating response in the Meeting Engagement module.}
    \label{tab:prompt_generate_response}
\end{table*}

\begin{table*}[htbp]
    \centering
    \footnotesize
    \begin{tabular}{p{15cm}}
    \toprule
    \#\# The output of response <Response>\\
    - The response should be a dictionary with the following format:
    \lstset{language=[LaTeX]TeX}
    \begin{lstlisting}
    {
        "thoughts": "<thoughts>",
        "speak": "<speak>"
    }
    \end{lstlisting}
    - <thoughts>: The reasoning or considerations to judge whether you can speak and decide what to say. At the beginning, you should state who you are representing. Then <thoughts> should explain what the latest utterance is about and then explain why you can or cannot speak. If you can speak, you should also explain how you decide what to say.\\
    - <speak>: The content you are going to speak. If you are not allowed to speak or you do not want to speak, the <speak> is empty.\\
    \\
    \#\# Example
    \lstset{language=[LaTeX]TeX}
    \begin{lstlisting}
    - Example 1:
    Below is an example of that pronunciations of Chinese names may be recognized as similar-sounding English words by the speech recognition system.
    <Person Name>
    'Sirui Zhao'
    <Attendees>
    [
        - 'San Zhang',
        - 'Si Li',
        - 'Sirui Zhao'
    ]
    <Meeting Transcript>
    Si Li: Good morning.
    Sirui Zhao: Hello!
    San Zhang: Hi!
    Si Li: OK, Let's start our meeting. There are still some people who haven't joined, so let's start first. Our topic today is the progress of environmental protection, three, do you have some thing to share on it?
    <Intents>
    [
        - 'The extent of plastic misuse'
    ]
    <Background>
    [
        {
            "Context":"Discussion about reducing air pollution",
            "Information":"The air pollution of our city is becoming serious. The goverment takes extreme measures to control the pollution by closing the factory and limiting the use private car."
        }
    ]
    <Response>
    {
        - "thoughts": "I'm representing Sirui Zhao in the meeting. In the last utterance, the appearance of 'three' is abrupt. Contextually, there is no need for numbers; phonetically, "Three" sounds like "Sirui.", which closely resembles 'Sirui' from the attendees, specifically <Person Name>. The speaker is most likely asking Sirui Zhao to share something on the progress of environment protection. So I need to give response. And based on the background information, I can share something about reduing air pollution.",
        - "speak": "Yes. The air pollution of our city is becoming serious. The goverment takes extreme measures to control the pollution by closing the factory and limiting the use private car."
    }
    \end{lstlisting}
    \textbf{CONTINUE ON THE NEXT PAGE} \\
    \bottomrule
    \end{tabular}
    \caption{Prompt used for generating response in the Meeting Engagement module (continued).}
\end{table*}

\begin{table*}[htbp]
    \centering
    \footnotesize
    \begin{tabular}{p{15cm}}
    \toprule
        \lstset{language=[LaTeX]TeX}
    \begin{lstlisting}
    - Example 2:
    Below is an example of that you should not speak since the latest utterance is a straightforward question or request or instructions to other participants.
    <Person Name>
    'Frank'
    <Attendees>
    [
        - 'John',
        - 'James',
        - 'Alice',
        - 'Bob'
        - 'Frank'
    ]
    <Meeting Transcript>
    Bob: James, that price is too high, we can not accept it.
    James: Ok, I will contact the supplier again and discuss the price.
    John: Thank you, James.
    John: OK, Let's go to the next topic. Alice, what is your progress on the project development?
    <Intents>
    [
        - 'Whether Bob fixed the bug I reported'
    ]
    <Background>
    [
        {
            "Context":"Report on the dataset preparation progress",
            "Information":'The dataset preparation is almost done. We are now working on the data cleaning and normalization. We expect to finish it by the end of the week.'
        }
    ]
    <Response>
    {
        - "thoughts": "I am representing Frank in the meeting. In the latest utterance, John is explicitly asking Alice about the project development. I can not speak.",
        - "speak": ""
    }
    \end{lstlisting}
    \#\# Note\\
    - You are representing <Person Name> in the meeting. You should respond to the cues from the attendees and the context of the meeting.\\
    - You should not interrupt others in the meeting.\\
    \bottomrule
    \end{tabular}
    \caption{Prompt used for generating response in the Meeting Engagement module (continued).}
\end{table*}

\begin{table*}[htbp]
    \centering
    \footnotesize
    \begin{tabular}{p{15cm}}
    \toprule
    - You are an Evaluation Agent responsible for assessing the response generated by a meeting AI assistant against the standard answer.\\
    - You are provided with the summary of the <StandardAnswer>.\\
    - You are provided with the raw <ActualResponse> generated by the meeting AI assistant.\\
    - Your task is to summarize the main points in the <ActualResponse>, and evaluate whether the main points in the <ActualResponse> match the main points in the <StandardAnswer>.\\
\\
    \#\# About the meeting AI Assistant\\
    - The meeting AI assistant is designed to represent the user to engage in a meeting.\\
\\
    \#\# About the <StandardAnswer>\\
    - The <StandardAnswer> is a list of strings that represents the main points of the ground truth response.\\
\\
    \#\# About the <ActualResponse>\\
    - The <ActualResponse> is a string that represents the response generated by the meeting AI assistant to the meeting content.\\
    - You should reference the <Transcript> to understand the context of the meeting and the information in the <ActualResponse>.\\
\\
    \#\# Guidelines for Evaluation\\
    - The evaluation process involves comparing the main points in the <StandardAnswer> and the <ActualResponse>.\\
    - Summarize the main points in the <ActualResponse> and keep the same granularity as the <StandardAnswer>.\\
    - The uninformative utterance about expressing goodness or politeness should not be considered as main points.\\
    ~~~~- For example, "If you need more help, please let me know." is not informative and should not be considered as a main point.\\
    - You should calculate a list that contains the index of the matching main points in the <ActualResponse> corresponding to the <StandardAnswer>. For example, if the first main point in the <ActualResponse> matches the second main point in the <StandardAnswer>, the first element of the list should be 2. And if the total number of main points in the <ActualResponse> is 1, the list should be [2].\\
    - Count the number of main points in the <ActualResponse> (ActualMainPointsCount).\\
    - Count the number of matching main points between the <ActualResponse> and the <StandardAnswer> (MatchingMainPointsCount).\\
    - The main points are considered matching if they are semantically similar.\\
\\
    \#\# Output Format\\
    - The output MUST be in the JSON format.\\
    - You MUST explain the process of evaluation before providing the evaluation results.\\
    - The output MUST include the following fields:\\
    ~~~~- Explanation: A explanantion of steps involved in the evaluation process. First, you should summarize the main points in the <ActualResponse>. Then, you should explain which main points in the <ActualResponse> match the main points in the <StandardAnswer> and mark the index of the matching main points in the <ActualResponse> corresponding to the <StandardAnswer> main points.\\
    ~~~~- ActualMainPoints: The list of main points in the <ActualResponse>.\\
    ~~~~- ActualMainPointsCount: The number of main points in the <ActualResponse>.\\
    ~~~~- MatchingMainPoints: The list of matching main points between the <ActualResponse> and the <StandardAnswer>.\\
    ~~~- MatchingIndex: The list of the index of the matching main points in the <ActualResponse> corresponding to the <StandardAnswer> main points. The length of the list should be the same as the ActualMainPointsCount. If ActualMainPointsCount is 5, the format of the list should be [1, 2, -1, -1, 4], which means the first, second, and fifth main points in the <ActualResponse> match the first, second, and fourth main points in the <StandardAnswer>. And the third and fourth main points in the <ActualResponse> do not match any main points in the <StandardAnswer>.\\
    ~~~~- MatchingMainPointsCount: The number of matching main points between the <ActualResponse> and the <StandardAnswer>.\\
    - If the <ActualResponse> is empty, the ActualMainPointsCount, MatchingMainPointsCount, RecallRate, and PrecisionRate should be 0.\\
    - Note that you must keep the length of the MatchingIndex the same as the ActualMainPointsCount, instead of the length of the <StandardAnswer>.
    \\
    \\
    \textbf{CONTINUE ON THE NEXT PAGE} \\
    \bottomrule
    \end{tabular}
    \caption{Prompt used for evaluating the generated response against the ground-truth one.}
    \label{tab:prompt_evaluation}
\end{table*}

\begin{table*}[htbp]
    \centering
    \footnotesize
    \begin{tabular}{p{15cm}}
    \toprule
    \lstset{language=[LaTeX]TeX}
    \begin{lstlisting}
    ## Example1
    <StandardAnswer>
    ["Calculated word error rate on Czech transcripts", "Conducted testing sessions with PERSON11 and PERSON18", "Contributed to the PROJECT3 deliverable", "Waiting for new tasks"]
    <ActualResponse>
    "Hi everyone. Over the past week, I calculated the word error rate on Czech transcripts using three versions of Czech ASR created by PERSON10. I also conducted a few testing sessions with PERSON11 and PERSON18, but they were not successful due to issues with the segmenters from ORGANIZATION1. I have updated the German transcripts in its corresponding path. I also backed up all the systems, including some new ones created today and last week. For the next week, I am waiting for new tasks. By the way, do we have the golden transcripts for the English videos?"
    <Evaluation>
    {
        "Explanation": "First, summarize the main points in the <ActualResponse>. The <ActualResponse> has the following main points: 1. Calculated word error rate on Czech transcripts. 2. Conducted testing sessions with PERSON11 and PERSON18. 3. Updated German transcripts in its corresponding path. 4. Backed up all the systems. 5. Waiting for new tasks. 6. Ask about the golden transcripts for the English videos. So, the number of main points in the <ActualResponse> is 6 and the length of the MatchingIndex is 6. Sencond, compare the main points in the <ActualResponse> with the main points in the <StandardAnswer>. The matching main points between the <ActualResponse> and the <StandardAnswer> are: 1. Calculated word error rate on Czech transcripts. This point maches with the first point in the <StandardAnswer>. 2. Conducted testing sessions with PERSON11 and PERSON18. This point matches with the second point in the <StandardAnswer>. 3. Waiting for new tasks. This point matches with the fourth point in the <StandardAnswer>. The number of matching main points between the <ActualResponse> and the <StandardAnswer> is 3. The other points in the <ActualResponse> do not match points in the <StandardAnswer>.",
        "ActualMainPoints": ["Calculated word error rate on Czech transcripts", "Conducted testing sessions with PERSON11 and PERSON18", "Updated German transcripts in its corresponding path", "Backed up all the systems", "Waiting for new tasks", "Ask about the golden transcripts for the English videos"],
        "ActualMainPointsCount": 6,
        "MatchingMainPoints": ["Calculated word error rate on Czech transcripts", "Conducted testing sessions with PERSON11 and PERSON18", "Waiting for new tasks"],
        "MatchingIndex": [1, 2, -1, -1, 4, -1],
        "MatchingMainPointsCount": 3
    }
    ## Example2
    <StandardAnswer>
    ["Confirm the task about writing a report about the calculation and share it with others", "Synthesize the information other team members have shared", "Wait for the next task"]
    <ActualResponse>
    "Sure, I will finish the calculation. I will also write a report about the calculation."
    <Evaluation>
    {
        "Explanation": "First, summarize the main points in the <ActualResponse>. The <ActualResponse> has the following main points: 1. Confirm the task about finishing the calculation and writing a report about it. The number of main points in the <ActualResponse> is 1 and the length of the MatchingIndex is 1. Sencond, compare the main points in the <ActualResponse> with the main points in the <StandardAnswer>. The matching main points between the <ActualResponse> and the <StandardAnswer> contains: 1. Confirm the task about finishing the calculation and writing a report about it. This point matches with the first point in the <StandardAnswer>. The number of matching main points between the <ActualResponse> and the <StandardAnswer> is 1. The other points in the <ActualResponse> do not match points in the <StandardAnswer>.",
        "ActualMainPoints": ["Confirm the task about finishing the calculation and writing a report about it"],
        "ActualMainPointsCount": 1,
        "MatchingMainPoints": ["Confirm the task about finishing the calculation and writing a report about it"],
        "MatchingIndex": [1],
        "MatchingMainPointsCount": 1,
    }
    \end{lstlisting}
    \\
    \bottomrule
    \end{tabular}
    \caption{Prompt used for evaluating the generated response against the ground-truth one (continued).}
\end{table*}

\begin{table*}[htbp]
    \centering
    \footnotesize
    \begin{tabular}{p{15cm}}
    \toprule
    - You are an Attribution Agent responsible for assessing the response generated by a meeting AI assistant and determining its source.\\
    - You are provided with the list of <ActualResponse>.\\
    - You are also provided with the transcript of the meeting content (<Transcript>) and the <ContextInfo> used to generate the <ActualResponse>.\\
    - Your task is to attribute the <ActualResponse> to the corresponding part of the <Transcript> or the <ContextInfo>.\\
    \\
    \#\# About the <ActualResponse>\\
    - The <ActualResponse> is a list that represents the response generated by the meeting AI assistant.\\
    \\
    \#\# About the <StandardResponse>\\
    - The <StandardResponse> is a list that represents the expected response.\\
    - The <StandardResponse> may be the same as <ActualResponse>, or it may not be.\\
    \\
    \#\# About the <Transcript>\\
    - The transcript are the collection of utterances from the meeting participants.\\
    - Each utterance is formatted as "Name: Content", where 'Name' is the speaker's name and 'Content' is their spoken text.\\
    - Utterances are in chronological order and may contain typos and grammatical errors.\\
    - The transcript ends at the time stamp when the meeting AI assistant should generate the response.\\
    - Example utterances:\\
        ~~~~PERSON1: Hello everyone, I'm glad to see you all here today. (id=0)\\
    \\
    \#\# About the <ContextInfo>\\
    - <ContextInfo> is a dictionary that contains <Intents> and <Background>.\\
    - <Intents> consists of the questions or topics that can generate the <ActualResponse>.\\
    - <Background> is a list of "Context" and "Information" pairs. For each pair, "Information" can be shared in the "Context" situation to generate the <ActualResponse>. And each pair can be used many times.\\
    \\
    \#\# Guidelines for Attribution\\
    - You need to decide whether the main points in the <ActualResponse> match the <StandardResponse>.\\
    - The number of  main points in the <ActualResponse> is not fixed. PointID is used to identify the main points in the <ActualResponse>.\\
    - When assessing whether the main points in the <ActualResponse> originate from the <Transcript> or the <ContextInfo>, consider the following:\\
    ~~~~1. If the main point has a similar or the same meaning as the <ContextInfo>. You should consider it as originating from the <ContextInfo>. \\
    ~~~~2. If the main point explicitly repeats or closely relates to any point already mentioned in the <Transcript>. However, casual interactions such as greetings or small talk are permissible and not regarded as sourced from the <Transcript>."\\
    - There are four situations for the origin of the main points in the <ActualResponse>:\\
    ~~~~1. The main point in the <ActualResponse> can originate from the <ContextInfo> but is not present in the <Transcript>. You should append [PointID, 1, 0] to the AttributionList.\\
    ~~~~2. The main point in the <ActualResponse> does not originate from the <ContextInfo> but originates from the <Transcript>. You should append [PointID, 0, 1] to the AttributionList.\\
    ~~~~3. The main point in the <ActualResponse> can originate from both the <ContextInfo> and the <Transcript>. You should append [PointID, 1, 1] to the AttributionList.\\
    ~~~~4. The main point in the <ActualResponse> does not originate from the <ContextInfo> and is not present in the <Transcript>. You should append [PointID, 0, 0] to the AttributionList.\\
    \\
    \#\# Output Format\\
    - The output MUST be in the JSON format.\\
    - You MUST explain the process of attribution for every main point in the <ActualResponse>.\\
    - Note that AttributionList should only contain the List of lists and should not contain any additional information or annotations.\\
    \\
    \textbf{CONTINUE ON THE NEXT PAGE} \\
    \bottomrule
    \end{tabular}
    \caption{Prompt used for the attribution of the generated response.}
    \label{tab:prompt_attribution}
\end{table*}

\begin{table*}[htbp]
    \centering
    \footnotesize
    \begin{tabular}{p{15cm}}
    \toprule
    - The output MUST include the following fields:\\
    ~~~~- Explanation: For every main point in the <ActualResponse>, explain the process of attribution. Especially, explain why the main point matches or does not match the <StandardResponse> and why it originates from the <Transcript> or the <ContextInfo>.\\
    ~~~~- AttributionList: A list of lists, where each list contains PointID and the attribution for a main point in the <ActualResponse>.\\
    ~~~~- PointsCount: The number of main points in the <ActualResponse>.\\
    \\
    \#\# Example
    \lstset{language=[LaTeX]TeX}
    \begin{lstlisting}
    - Example 1:
    <Transcript>
    PERSON13: Hi. Hello [PERSON6]. Hello [PERSON19]. Thanks for, uhm. (id=0)
    PERSON6: Hi everyone. (id=1)
    PERSON19: Hi. (id=2)
    PERSON13: Yeah, great. Thanks for joining and, uh, yeah okay. So, yeah. Uh, I I see that people have written up ehm what they did. (id=3)
    PERSON19: Hi [PERSON13], I can hear you. (id=4)
    PERSON13: Yeah. [PROJECT3] deliverables. So, I'll try to provide the links-. Or those who of you, who are already working on the deliverables, please mention that. And yeah. Let's let's go quickly over what what have done. So [PERSON6] you are the first on the list. Ehm, ehm, so please briefly update what what you have been working on. And what what is the plan for the next week. (id=5)

    <ContextInfo>
    {
        "Intents": [
            "What [PERSON6] has been working on and the plan for the next week?"
        ],
        "Background": [
            {
                "Context": "Update on recent work and plans for the next week",
                "Information": "This week I had fewer tasks. I calculated the word error rate on Czech transcripts using three versions of Czech ASR created by [PERSON10]. There were significant mismatches between the golden transcript and its corresponding video. I conducted testing sessions with [PERSON11] and [PERSON18], which were not successful due to issues with segmenters from [ORGANIZATION1]. I also contributed to the [PROJECT3] deliverable for the punctuator and through caser."
            }
        ]
    }

    <StandardResponse>
    ["I calculated the word error rate on Czech transcripts"]

    <ActualResponse>
    ["Calculated word error rate on Czech transcripts", "Conducted testing sessions with PERSON11 and PERSON18"]
    
    <Evaluation>
    {
        "Explanation": "1. Calculated word error rate on Czech transcripts. This point matches the standard response. "I calculated the word error rate on Czech transcripts" is present in the ContextInfo. Therefore, the attribution is [1, 1, 0]. 2. Conducted testing sessions with PERSON11 and PERSON18. The point does not match the standard response. "I conducted testing sessions with [PERSON11] and [PERSON18]" is present in the BackgroundKnowledge. Therefore, the attribution is [2, 1, 0].",
        "AttributionList": [[1, 1, 0], [2, 1, 0]],
        "PointsCount": 2
    }
    \end{lstlisting}
    \\
    \bottomrule
    \end{tabular}
    \caption{Prompt used for the attribution of the generated response (continued).}
\end{table*}

\begin{table*}[htbp]
    \centering
    \footnotesize
    \begin{tabular}{p{15cm}}
    \toprule
  - Your task is to update the summary of the utterances of \{participant\} in a meeting transcript.\\
  - You are provided with a <Transcript Snippet> that contains a portion of the meeting transcript.\\
  - You are also provided with <Previous Summary> which contains the summary of utterances for \{participant\} in other parts of the meeting.\\
  - You need to update the <Previous Summary> based on the utterances of the \{participant\} in the <Transcript Snippet>.\\
\\
  \#\# On the provided <Transcript Snippet>\\
  - Transcripts are the collection of utterances from the meeting participants.\\
  - The transcript data is deidentified. Speakers and other named entities are not identified by names, but rather by IDs in the format ENTITYNUMBER (e.g. PERSON1 or PROJECT3) or just ENTITY (e.g. PATH). \\
  - Speaker IDs at the beginning of transcript lines are enclosed in round brackets, all other deidentified entities in square brackets.\\
  - Each utterence ends with "(id=x)", which is the utterance id, an increasing number from 0 to indicate the serial number of utterance in the whole meeting transcript.\\
  - The provided transcript snippet maybe not start from the beginning of the meeting.\\
  - Example utterances:\\
    ~~~~(PERSON1) Hello everyone, I'm glad to see you all here today. (id=0)\\
    ~~~~(PERSON2) Hi, I'm excited to be here. (id=1)\\
    ~~~~(PERSON3) I'm looking forward to the discussion. [PERSON1] mentioned that the project is going well. (id=2)\\
\\
  \#\# On the <Previous Summary>\\
  - The <Previous Summary> is a structured summary of the utterances of \{participant\} in the meeting transcript.\\
  - The <Previous Summary> contains two parts, "wanted information" and "provided information".\\
    ~~~~- "wanted information" is a list of questions made by the \{participant\}.\\
    ~~~~- "provided information" is the information provided by the {participant} to others. It is a list of Context and Information pairs, where the "Context" is the context in which where the \{participant\} provides the "Information".\\
\\
  \#\# Instructions on updating the <Previous Summary>\\
  - Identify the utterances of \{participant\} in the <Transcript Snippet>.\\
  - If \{participant\} does not speak in the <Transcript Snippet>, do NOT update the <Previous Summary>.\\
  - Focus on only the informative utterances and ignore the greetings, appreciation, simple acknowledge and other chit chat.\\
  - Extract the "wanted information" and "provided information" from the <Transcript Snippet>.\\
  - You should try to use original utterances as much as possible after removing noise words and polishing them for better readability.\\
  - The second or third personal pronoun (you, he, she, they) in the utterances should be properly replaced with the corresponding participant's ID to avoid ambiguity.\\
  - Use the extracted information to update the <Previous Summary>.\\
  - You can modify the existing "wanted information" and "provided information" or add new information, but do not remove any existing information.\\
  - You MUST NOT mix the information provided by \{participant\} and other participants while updating the <Previous Summary>.\\
  - You MUST NOT miss any important information provided by \{participant\} in the <Transcript Snippet>.\\
\\
  \#\# Requirement on the output format\\
  - You MUST explain your thoughts and steps of updating the <Previous Summary> before providing the updated summary.\\
  - Output must be in Json format with the "Thoughts" and "Updated Summary" as the key.\\
    - The "Thoughts" is your thoughts and steps of updating the <Previous Summary>.\\
    - The "Updated Summary" contains the updated summary of the utterances for \{participant\}.\\
    \\
    \textbf{CONTINUE ON THE NEXT PAGE} \\
    \bottomrule
    \end{tabular}
    \caption{Prompt used for extracting context information.}
    \label{tab:prompt_context_extraction}
\end{table*}

\begin{table*}[htbp]
    \centering
    \footnotesize
    \begin{tabular}{p{15cm}}
    \toprule
    \lstset{language=[LaTeX]TeX}
    \begin{lstlisting}
      ## Example 1
      - Here is an example of updating the utterances summary for PERSON2. You can refer to this example for better understanding.
      - Suppose the transcript snippet contains the following utterances:
          (PERSON3) Good moring. (id=2)
          (PERSON1) Let's get started with today's meeting on the recent progress of our software development project. We'll go through updates from each team and discuss any blockers or issues. [PERSON2], could you start with the development updates?". (id=3)
          (PERSON2) Sure, [PERSON1]. We've made significant progress this sprint. We completed the implementation of the new authentication module and integrated it with our existing systems. (id=4)
          (PERSON1) That's great to hear, [PERSON2]. How about the feature for real-time notifications? Is it on track? (id=5)
          (PERSON2) Yes, it is. We're about 75% done with it. The core functionality is in place, and we are now working on optimizing the delivery speed and ensuring it works seamlessly across different devices. (id=6)
      - Suppose the previous summary of PERSON2 contains the following information:
          {{
            "wanted information": [],
            "provided information": []
          }}
      - The thoughts and updated summary will be:
          {{
            "Thoughts":"In the transcript snippet, PERSON2 responds to PERSON1's questions about the development updates and the progress of the feature for real-time notifications. This information can be added to the "provided information" for PERSON2."
            "Updated Summary": 
              {{
                "wanted information": [],
                "provided information": [
                  {{
                    "Context": "Respond to other participant's question about the development updates",
                    "Information": "We've made significant progress this sprint. We completed the implementation of the new authentication module and integrated it with our existing systems."
                  }},
                  {{
                    "Context": "Respond to other participant's question about the progress of the feature for real-time notifications",
                    "Information": "We're about 75% done with it. The core functionality is in place, and we are now working on optimizing the delivery speed and ensuring it works seamlessly across different devices."
                  }}
                ]
              }}
          }}
    
      ## Example 2
      - Here is another example of updating the utterances summary for PERSON2. You can refer to this example for better understanding.
      - Suppose the transcript snippet contains the following utterances:
          (PERSON3) Good moring. (id=2)
          (PERSON1) Let's get started with today's meeting on the recent progress of our software development project. We'll go through updates from each team and discuss any blockers or issues. [PERSON2], could you start with the development updates?". (id=3)
          (PERSON2) Sure, [PERSON1]. We've made significant progress this sprint. We completed the implementation of the new authentication module and integrated it with our existing systems. (id=4)
    \end{lstlisting}
    \textbf{CONTINUE ON THE NEXT PAGE} \\
    \bottomrule
    \end{tabular}
    \caption{Prompt used for extracting context information (continued).}
\end{table*}

\begin{table*}[htbp]
    \centering
    \footnotesize
    \begin{tabular}{p{15cm}}
    \toprule
    \lstset{language=[LaTeX]TeX}
    \begin{lstlisting}
          (PERSON1) That's great to hear, [PERSON2]. How about the feature for real-time notifications? Is it on track? (id=5)
          (PERSON2) Yes, it is. We're about 75% done with it. The core functionality is in place, and we are now working on optimizing the delivery speed and ensuring it works seamlessly across different devices. (id=6)
          (PERSON3) [PERSON2], have you had a chance to address the bug I reported last week related to the authentication module? (id=7)
          (PERSON2) Yes, [PERSON3]. We identified the root cause of the bug, and it's been fixed. It was due to a conflict with a third-party library we were using. (id=8)
          (PERSON3) That's good to hear. Thank you, [PERSON2]. (id=9)
          (PERSON1) [PERSON2], for the next step of the project, I'd like you first complete the real-time notifications feature, and then focus on the chatbot development. (id=10)
          (PERSON2) Understood, I will do that. (id=11)
          (PERSON2) By the way, what's the timeline of our project? (id=12)
          (PERSON1) We are aiming to finish the project by the end of August. (id=13)
          (PERSON2) Ok, I know. (id=14)
          (PERSON1) Let's move to the next topic. [PERSON3], could you provide an update on the testing progress? (id=15)
          (PERSON3) Sure. Certainly. We've conducted tests on the new authentication module, and everything looks good so far. (id=16)
          (PERSON1) Mhm. (id=17)
          (PERSON3) We are now preparing for the testing of the real-time notifications feature. (id=18) 
          (PERSON2) In our development process, we accumulated some test cases which may help you. (id=19)
          (PERSON3) That's helpful, thank you. (id=20)
      - Suppose the previous summary of PERSON2 contains the following information:
          {{
            "wanted information": [],
            "provided information": [
              {{
                "Context": "Respond to other participant's question about the development updates",
                "Information": "We've made significant progress this sprint. We completed the implementation of the new authentication module and integrated it with our existing systems."
              }},
              {{
                "Context": "Respond to other participant's question about the progress of the feature for real-time notifications",
                "Information": "We're about 75% done with it. The core functionality is in place, and we are now working on optimizing the delivery speed and ensuring it works seamlessly across different devices."
              }}
            ]
          }}
      - The thoughts and updated summary will be:
          {{
            "Thoughts":"In the transcript snippet, the dicussion between PERSON1 and PERSON2 about the progress of the development and the feature for real-time notifications are already included in the previous summary. PERSON2 responds to PERSON3's question about the bug in the authentication module, which can be added to the "provided information" for PERSON2. PERSON2 asks about the timeline of the project, which can be added to the "wanted information" for PERSON2. PERSON2 also comments on PERSON3's statement about the testing progress, offering to provide some test cases, which can be added to the "provided information" for PERSON2."
            "Updated Summary":
              {{
                "wanted information": [
                  "What's the timeline of the project?"
                ],
                "provided information": [
                  {{
                    "Context": "Respond to other participant's question about the bug in the authentication module",
    \end{lstlisting}
    \textbf{CONTINUE ON THE NEXT PAGE} \\
    \bottomrule
    \end{tabular}
    \caption{Prompt used for extracting context information (continued).}
\end{table*}

\begin{table*}[htbp]
    \centering
    \footnotesize
    \begin{tabular}{p{15cm}}
    \toprule
    \lstset{language=[LaTeX]TeX}
    \begin{lstlisting}

                    "Information": "We identified the root cause of the bug reported by [PERSON3], and it's been fixed. It was due to a conflict with a third-party library we were using."
                  }},
                  {{
                    "Context": "Comment on other participant's statement about the testing progress",
                    "Information": "In our development process, we accumulated some test cases which are helpful to testing."
                  }}
                ]
              }}
          }}

      ## Example 3
      - Here is an example of updating the utterances summary for PERSON6. You can refer to this example for better understanding.
      - Suppose the transcript snippet contains the following utterances:
          (PERSON13) Hi.
          Hello [PERSON6].
          Hello [PERSON19].
          Thanks for, uhm. (id=0)
          (PERSON6) Hi everyone. (id=1)
          (PERSON19) Hi. (id=2)
          (PERSON13) Yeah, great.
          Thanks for joining and, uh, yeah okay.
          So, yeah.
          Uh, I I see that people have written up ehm what they did. (id=3)
          (PERSON19) Hi [PERSON13], I can hear you. (id=4)
          (PERSON13) Yep, that's great.
          Uh, and also you were evaluating-.
          Yes, so that's that's re re record.
          What you did.
          So what I have, uh, on my mind now is uh, uh, well, uh, preparations.
          So, uh, [PERSON13], uh I am busy, uh, with the IW SLT, uh, write-up.
          Uh, that was the, uh, the wra last part that I did.
          Now busy with interviewing people people to uh to replace those who are em moving forward <laugh/> so to say.
          So there is number of colleagues on projects that I am supervising, uh, that who are going for studies abroad and other things.
          Uh, so, uh, what I think we should focus on is the demo for Project Officer.
          Then we need to focus on the ladder climbing, uh, which is building uh, uh, [PROJECT3] test set plus, uh, regularly, uh, testing on it.
          Ehm, and, ehm what else, uh, the deliverables.
          Yeah.
          [PROJECT3] deliverables.
          So, I'll try to provide the links-.
          Or those who of you, who are already working on the deliverables, please mention that.
          And yeah.
          Let's let's go quickly over what what have done.
          So [PERSON6] you are the first on the list.
          Ehm, ehm, so please briefly update what what you have been working on.
          And what what is the plan for the next week. (id=5)
          (PERSON6) <other_noise/>
          So, luckily. 
          <laugh/>
          Not luckily but this week I had like quite less tasks to do.
          So first I calculated the word error rate on Czech transcripts using that three versions of, uh, Czech ASR which [PERSON10] created.
          And so yesterday [PERSON10] told me that they were, uh, and the golden transcript and its corresponding video there were there were huge huge mismatch.
          And I <unintelligible/> and he said to update me.
    \end{lstlisting}
    \textbf{CONTINUE ON THE NEXT PAGE} \\
    \bottomrule
    \end{tabular}
    \caption{Prompt used for extracting context information (continued).}
\end{table*}

\begin{table*}[htbp]
    \centering
    \footnotesize
    \begin{tabular}{p{15cm}}
    \toprule
    \lstset{language=[LaTeX]TeX}
    \begin{lstlisting}
          And then we conducted a few testing sessions with [PERSON11] and [PERSON18].
          And they were not quite successful because the segmenters from, uh, uh, [ORGANIZATION1] they were still down and [PERSON12] today he is he is working on them. (id=6)
          (PERSON13) Mhm. (id=7)
          (PERSON6) And lastly yeah I think I did-.
          Uh, it was the input in the [PROJECT3] deliverable of for the punctuator and through caser.
          <unintelligible/> (id=8)
          (PERSON13) Mhm, yeah. (id=9)
          (PERSON6) And I don't have-.
          I think that apart from the testing sessions to do this week so I am waiting for new tasks. (id=10)
          (PERSON13) Yeah, so.
          So the word error rate, there is also the English, uh, transcripts?
          Ehm, and also we should have from [PERSON9] the German one, right?
          So. (id=11)
          (PERSON6) Yeah, yeah, yeah. (id=12)
          (PERSON13) So, so I will make it to do-. (id=13)
          (PERSON6) So I have updated the German transcripts in its corresponding path and like do we have the golden transcripts for the English videos? (id=14)
          (PERSON13) Yes, that's the other part.
          Because this is the consecutively translated videos.
          So there is always the English speaker and then the Czech speaker who repeats the same content.
          And [PERSON7] has split the video and while the English part should be more reliable, uh, the Czech part has been done simply by using the other ends.
          So the Czech video has been cut using the English time stamps.
          Like the end of English and the beginning of the next English segment.
          Uh, so it's like like interleave the the the other way round.
          So that's why I'm not surprised that the Czech video is, uh, imprecise in timing.
          But still, I was not expecting it to be that bad.
          So, uh, that is something that, yeah.
          [PERSON10], can you maybe tell us more details about that? (id=15)
          (PERSON10) Yeah, yeah.
          Well, I just like listened to the audio and followed the talk transcript <other_noise/>
          and it was completely off.
          I think it is-.
          There must be some miss-match because-. (id=16)
          (PERSON13) Mhm. (id=17)
          (PERSON10) Yeah, yeah. 
          The transcription is for the completely different audio than it's in the subdirectory. (id=18)
          (PERSON6) Mhm. (id=19)
          (PERSON13) Oh, so then someone must have like messed it up. (id=20)
          (PERSON10) Yeah, yeah, I I may maybe it's just like uh.
          Maybe the files are just switched between the subdirectories? (id=21)
          (PERSON13) Mhm. (id=22)
          (PERSON10) I I haven't checked but-.
          Uh, yeah, there is some some serious mismatch there. (id=23)
          (PERSON13) Yeah, so [PERSON10] can you coul could you do this check?
          It should not be hard
          Like try listening to all the files that are within this demo for [PERSON15], uh, and try to locate the correct file, the appropriate files.
          But we should have, we should have the transcripts ready for all of those.
          So we should be able to, uh, to evaluate it.
          And also for the English ones we have the translations.
          So for the English ones [PERSON6], uh, I would like you to evaluate not only the word error rate of the ASR.
          But also the machine translation quality or at the SLT even.
          Uh, with the translation quality into German and Czech.
          Both are available. (id=24)
          (PERSON6) Okay. (id=25)

    \end{lstlisting}
    \textbf{CONTINUE ON THE NEXT PAGE} \\
    \bottomrule
    \end{tabular}
    \caption{Prompt used for extracting context information (continued).}
\end{table*}

\begin{table*}[htbp]
    \centering
    \footnotesize
    \begin{tabular}{p{15cm}}
    \toprule
    \lstset{language=[LaTeX]TeX}
    \begin{lstlisting}
          (PERSON13) We have these files ready. (id=26)
          (PERSON6) And so these, for like a German audio and English [PROJECT1]. (id=27)
          (PERSON13) English, uh, input for English sound.
          We have the golden English transcript, so you can check the ASR.
          And we also have the translation into Czech and into German.
          So you can also evaluate directly the translation quality, uh, of that. (id=28)
          (PERSON6) Okay, yeah. (id=29)
          (PERSON13) Yeah, so this is, this is an important, uh, task, uh, to do, uh wr also for German and English audios.
          And another to do, uh, bleu or SLTF, uh, for, uh, German and Czech translations of, uh, English. (id=30)
      - Suppose the previous summary of PERSON6 contains the following information:
          {{
            "wanted information": [],
            "provided information": []
          }}
      - The thoughts and updated summary will be:
          {{
            "Thoughts":"PERSON6 responds to PERSON13's questions about the work done and the plan for the next week, which can be added to the 'provided information' for PERSON6. PERSON6 responds to PERSON13's questions about the word error rate for the English and German transcripts, which can be added to the 'provided information' for PERSON6. PERSON6 also asks about the golden transcripts for the English videos, which can be added to the 'wanted information' for PERSON6."
            "Updated Summary": 
              {{
                "wanted information": [
                  "Do we have the golden transcripts for the English videos?"
                ],
                "provided information": "wanted information": [
                  "Do we have the golden transcripts for the English videos?"
                ],
                "provided information": [
                  {{
                    "Context": "Respond to other participant's question about the work done and the plan for the next week",
                    "Information": "This week I had quite less tasks to do. So first I calculated the word error rate on Czech transcripts using three versions of Czech ASR created by [PERSON10]. And so yesterday [PERSON10] told me that there were huge huge mismatch between the golden transcript and its corresponding video. And he said to update me. And then we conducted a few testing sessions with [PERSON11] and [PERSON18]. And they were not quite successful because the segmenters from [ORGANIZATION1] were still down and [PERSON12] today is working on them. And lastly, I think I did the input in the [PROJECT3] deliverable for the punctuator and through caser. Apart from the testing sessions to do this week so I am waiting for new tasks."
                  }},
                  {{
                    "Context": "Respond to other participant's question about the word error rate for the English and German transcripts",
                    "Information": "I have updated the German transcripts in its corresponding path, and I don't konw if we have the golden transcripts for the English videos."
                  }}
                ]
              }}
          }}
      
      ## Note
      - You MUST follow the instructions and examples provided.
      - Similar to examples above, you should try to use original utterances as much as possible after removing noise words and polishing them for better readability.
      - You MUST NOT put the information provided by other participants or questions of other participants in the updated summary of {participant}.
      - You MUST NOT miss any important information provided by {participant} in the <Transcript Snippet>.
      - You MUST give the output in the required format.
    \end{lstlisting}
    \\
    \bottomrule
    \end{tabular}
    \caption{Prompt used for extracting context information (continued).}
\end{table*}

\begin{table*}[htbp]
    \centering
    \footnotesize
    \begin{tabular}{p{15cm}}
    \toprule
    - You are an NLP expert agent tasked with generating an evaluation dataset to assess \{person\_id\}'s response abilities in the categories of 'Chime In', 'Explicit Cue', 'Implicit Cue', based on the provided transcript.\\
    - The conversation may involve multiple speakers, but your focus should solely be on \{person\_id\}.\\
    - Given the transcript contains lengthy utterances, selectively include only the highest quality exchanges in the evaluation dataset.\\
    - Exclude chit-chat or unmeaningful utterances such as ["emm," "okay," "mhm," "uh-huh," "yeah," "oh," "right," "hmm"] from the evaluation dataset.\\
    - Ensure that \{person\_id\}'s responses are substantive and meaningful. Exclude responses from \{person\_id\} that are simple acknowledgments or confirmations like "Yeah, yeah, definitely, yeah" or "Okay."\\
    - The transcript data is deidentified. Speakers and other named entities are identified by IDs in the format ENTITYNUMBER (e.g., PERSON1, Speaker1 or PROJECT3) or simply as ENTITY (e.g., PATH).\\
    \\
    \#\# Evaluation Type\\
    - Chime In: When \{person\_id\} spontaneously contributes to the conversation without being directly prompted.\\
    - Usually Chime In is when \{person\_id\} is not already engaged in the conversation but chimes in with a relevant comment or question.  \\
    - Explicit Cue: When \{person\_id\}'s name is specifically mentioned by another Speaker in utterance with ID, then \{person\_id\} responds to a clear and direct question or prompt towards \{person\_id\}.\\
    - Implicit Cue: When \{person\_id\}'s name is not specifically mentioned by Speaker in utterance with ID, but \{person\_id\} responds to a less direct prompt or follows up on information that suggests a response is needed.\\
    - Usually Implicit Cue is when \{person\_id\} is already engaged in the conversation and responds to a follow-up question from Speaker in utterance with ID.\\
    \\
    \#\# Output Format\\
    - Output must be in Json format. Here is the skeleton of the output format with explanation:\\
    - Explanation: Your reason for selecting the evaluation instance and for categorizing it.\\
    - Type: The category of the evaluation instance: 'Chime In', 'Explicit Cue', 'Implicit Cue'.\\
    - Response IDs: The id or ids of the \{person\_id\}'s response from the transcript. Include all Response IDs that are relevant to the evaluation instance. If there are multiple Response IDs, separate them with commas.\\
    - ID: The utterance id that \{person\_id\} responds to.\\
    - Speaker: The speaker of the utterance with the ID.\\
    - Maintain the chronological order of the transcript when generating the evaluation dataset. ID MUST precede Response IDs.\\
    - Response IDs must be from \{person\_id\}'s responses only and ID must be from the speaker's utterance that \{person\_id\} responds to.\\
    - Please return all suitable evaluation instances in the transcript. If you don't find any suitable instances for a category, you can leave the evaluation dataset empty. Please ensure you have thought through the transcript carefully before leaving the evaluation dataset empty.\\
    \\
    \#\# Example: Below are two examples of transcript and the corresponding evaluation datasets generated to assess PERSON18's response abilities. You can refer to these examples when generating \{person\_id\}'s evaluation dataset.\\
    \\
    <Transcript>\\
    "speaker": "Speaker 19", "content": "If you want, I can resend it again. (id=71)"\\
    "speaker": "Speaker 13", "content": "Space tokeniser. <unintelligible/> Yes, so es essentially to answer your question in the email. We have to switch to and we have for  the IWSLT. We have to switch to SacreBLEU and SacreBLEU does its own tokenisation before scoring. So there is no-. Let's let's simply forget NLTK bleu score. That is not reliable. (id=72)"\\
    "speaker": "Speaker 18", "content": "Yes, but-. (id=73)"\\
    "speaker": "Speaker 19", "content": "Yes, but we can combine our tokeniser with NLTK. (id=74)"\\
    "speaker": "Speaker 13", "content": "Uf. Let's not do that. Let's just forget it. Let's let's just use SacreBLEU. (id=75)"\\
    "speaker": "Speaker 19", "content": "Okay. (id=76)"\\
    "speaker": "Speaker 18", "content": "I I have one comment about it. (id=77)"\\
    "speaker": "Speaker 13", "content": "Mhm.  Yeah. (id=78)"\\
    "speaker": "Speaker 18", "content": "You sh should use tokeniser before enverse segmenter. (id=79)"\\
    "speaker": "Speaker 13", "content": "Yes, that's it. Yeah. (id=80)"\\
    "speaker": "Speaker 18", "content": "Because it's much better. Because it can rely on the on the dots and commas and question marks and so on. And you can you can check my script which does tokeniser, enverse segmenter and then de-tokeniser. And here is the path in the document.And-. (id=81)"\\
    "speaker": "Speaker 13", "content": "Yeah. (id=82)"\\
    "speaker": "Speaker 18", "content": "And it's it's using the Moses seg tokeniser and detokeniser. And it needs the the language tag as the first argument and then reference. (id=83)"\\
    "speaker": "Speaker 13", "content": "Yeah. So [PERSON19], do you do you fo-? Do you understand? (id=84)"\\
    "speaker": "Speaker 19", "content": "Yeah. (id=85)"\\

    \textbf{CONTINUE ON THE NEXT PAGE} \\
    \bottomrule
    \end{tabular}
    \caption{Prompt used for extracting test cases from meeting transcript.}
    \label{tab:prompt_case_extraction}
\end{table*}

\begin{table*}[htbp]
    \centering
    \footnotesize
    \begin{tabular}{p{15cm}}
    \toprule
    \lstset{language=[LaTeX]TeX}
    \begin{lstlisting}
    <Evaluation Dataset>
    [
        "Explanation": "PERSON18's utterance is informative and suitable for evaluation dataset. PERSON18 spontaneously contributes to the conversation without being directly prompted. This is a Chime In instead of Implicit Cue since [PERSON18] is not already engaged in the conversation.",
        "Type": "Chime in",
        "Response ID": [77, 79, 81, 83]
        "ID": 76,
        "Speaker": "Speaker 19",
    ]
    \end{lstlisting}    
    \#\# Example 2\\ 
   <Transcript>
   
   "speaker": "Speaker 13", "content": "Yes, so I need to review these and the internal deadline is in 6 days from now. Uh, so, hopefully I will get back to all of you. To each of you independently towards the end of the week if there is anything unclear. So that we meet the internal deadline on the 8th. Yeah, okay. Great. Uh. So [PERSON18], what what is your progress? (id=117)"\\
   "speaker": "Speaker 18", "content": "Hmhm. Yes, and by reading the papers I found an interesting tool. (id=118)"\\
   "speaker": "Speaker 13", "content": "Mhm. (id=119)"\\
   "speaker": "Speaker 18", "content": "I found out that it's possible to measure out the speech rate by cutting the syllables. And there is one tool. One patent tool, which can detect the gender of speaker and the speech rate. (id=120)"\\
   "speaker": "Speaker 13", "content": "Mhm. (id=121)"\\
   "speaker": "Speaker 18", "content": "And some other characteristics. So we can try it and make a dashboard out of it. (id=122)"\\
   "speaker": "Speaker 13", "content": "Mhm. That's that's useful thing. Uh, and later on we could even create models like-. If we if we recognise that someone is speaking too fast, we could use like a harsher summarisation. (id=123)"\\
   "speaker": "Speaker 18", "content": "Yes. (id=124)"\\
   "speaker": "Speaker 13", "content": "So we could be reducing reducing their speech mole with a different model. (id=125)"\\
   "speaker": "Speaker 18", "content": "Yes, and there was also speech modes. Like whether it was angry or normal and so on. (id=126)"\\
   "speaker": "Speaker 13", "content": "Mhm. (id=127)"\\
   "speaker": "Speaker 18", "content": "But I have no idea how the tool works in practice. I I I I saw it only in Gi GitHub and I buy it. (id=128)"\\
   "speaker": "Speaker 13", "content": "Yeah, uh. (id=129)"\\
   "speaker": "Speaker 18", "content": "So we can try it and make a dashboard out of it. (id=130)"\\
   \begin{lstlisting}
   <Evaluation Dataset>
   [ 
     "Explanation": "PERSON18's utterance is informative and suitable for evaluation dataset. PERSON18 was directly prompted by Speaker 13. This is an Explicit Cue.",
     "Type": "Explicit Cue",
     "Response ID": [118, 120, 122, 124, 126, 128, 130]
     "ID": 117,
     "Speaker": "Speaker 13",
   ]   
   \end{lstlisting}
  - Please refer to the examples provided to ensure consistency and coherence in generating the evaluation dataset. The evaluation dataset must be in json format.\\
    \\

    \bottomrule
    \end{tabular}
    \caption{Prompt used for extracting test cases from meeting transcript (continued).}
\end{table*}

\end{document}